\documentclass{article}
\usepackage[preprint]{spconf}

\usepackage{multirow}
\usepackage{graphicx}
\usepackage{amsmath}
\usepackage[psamsfonts]{amssymb}
\usepackage{amsxtra}
\usepackage{threeparttable}
\usepackage{amsfonts}
\usepackage{mathrsfs}
\usepackage[caption=true, font=footnotesize,
nearskip=0.0pt,farskip=0.0pt, captionskip=0.0pt]{subfig}
\usepackage{arydshln}
\usepackage{cite}
\usepackage{enumitem}

\makeatletter
\renewcommand\section{\@startsection{section}{1}{\z@}
                      {0.5ex \@plus 0ex \@minus -2ex}
                      {0.5ex \@plus 0ex}
                      {\normalfont\Large\bfseries}}
\renewcommand\subsection{\@startsection{subsection}{2}{\z@}
                      {0.5ex \@plus 0ex \@minus -2ex}
                      {0.5ex \@plus 0ex}
                      {\normalfont\large\bfseries}}
\renewcommand\subsubsection{\@startsection{subsubsection}{3}{\z@}
                      {0.5ex \@plus 0ex \@minus -2ex}
                      {0.5ex \@plus 0ex}
                      {\normalfont\normalsize\bfseries}}
\def\@listi{\leftmargin\leftmargini
            \parsep 1.0pt
            \topsep 0.2\baselineskip \@minus 0.1\baselineskip
            \itemsep 1.0pt \relax}
\let\@listI\@listi
\makeatother

\AtBeginDocument{
  \abovedisplayskip=0.3\abovedisplayskip
  \abovedisplayshortskip=0.3\abovedisplayshortskip
  \belowdisplayskip=0.3\belowdisplayskip
  \belowdisplayshortskip=0.3\belowdisplayshortskip
}

\title{Automatic Exposure Compensation for Multi-Exposure Image Fusion}
\name{Yuma Kinoshita \qquad Sayaka Shiota \qquad Hitoshi Kiya}
\address{Tokyo Metropolitan University, Tokyo, Japan}
\copyrightnotice{\copyright\ IEEE 2018}
\toappear{To appear in {\it Proc.\ ICIP2018 October 07-10, 2018, Athens, Greece}}
\begin{document}\sloppy
\setlength{\parskip}{0.0pt}
\setlength{\tabcolsep}{1.0pt}
\setlength{\textfloatsep}{1.0pt}
\setlength{\floatsep}{0.0pt}
\setlength{\abovecaptionskip}{0.0pt}
\setlength{\belowcaptionskip}{1.0pt}
\setlength{\intextsep}{0.0pt}
\setlength{\dblfloatsep}{0.0pt}
\setlength{\dbltextfloatsep}{2.0pt}
\ninept
\maketitle
\begin{abstract}
  This paper proposes a novel luminance adjustment method based on 
  automatic exposure compensation for multi-exposure image fusion.
  Multi-exposure image fusion is a method to produce images without saturation
  regions, by using photos with different exposures.
  In conventional works, it has been pointed out that
  the quality of those multi-exposure images can be improved by adjusting
  the luminance of them.
  However, how to determine the degree of adjustment has never been discussed.
  This paper therefore proposes a way to automatically determines the degree
  on the basis of the luminance distribution of input multi-exposure images.
  Moreover, new weights, called ``simple weights'', for image fusion
  are also considered for the proposed luminance adjustment method.
  Experimental results show that the multi-exposure images adjusted by the proposed method
  have better quality than the input multi-exposure ones in terms of the well-exposedness.
  It is also confirmed that
  the proposed simple weights provide the highest score of
  statistical naturalness and discrete entropy in all fusion methods.
\end{abstract}
\renewcommand{\thefootnote}{\fnsymbol{footnote}}
\footnote[0]{
  This work was supported by JSPS KAKENHI Grant Number JP18J20326.}
\renewcommand{\thefootnote}{\arabic{footnote}}
\section{Introduction}
  The low dynamic range (LDR) of the imaging sensors used in modern digital cameras
  is a major factor preventing cameras from capturing images as good as those with human vision.
  Various methods for improving the quality of a single LDR image by enhancing the contrast
  have been proposed\cite{zuiderveld1994contrast, wu2017contrast, kinoshita2017pseudo}.
  However, contrast enhancement cannot restore saturated pixel values in LDR images.
  
  Because of such a situation, the interest of multi-exposure image fusion has
  recently been increasing.
  Various research works on multi-exposure image fusion have so far been reported
  \cite{goshtasby2005fusion,mertens2009exposure,saleem2012image,wang2015exposure,
  li2014selectively,sakai2015hybrid, nejati2017fast, prabhakar2017deepfuse}.
  These fusion methods utilize a set of differently exposed images,
  ``multi-exposure images'', and fuse them to produce an image with high quality.
  Their development was inspired by high dynamic range (HDR) imaging techniques
  \cite{debevec1997recovering,reinhard2002photographic,oh2015robust,
  kinoshita2016remapping,kinoshita2017fast,kinoshita2017fast_trans,huo2016single,
  murofushi2013integer,murofushi2014integer,dobashi2014fixed}.
  The advantage of these methods, compared with HDR imaging techniques, is that
  they can eliminate three operations:
  generating HDR images, calibrating a camera response function (CRF),
  and preserving the exposure value of each photograph.

  However, the conventional multi-exposure image fusion methods have several problems
  due to the use of a set of differently exposed images.
  The set should consist of a properly exposed image,
  overexposed images and underexposed images,
  but determining appropriate exposure values is problematic.
  Moreover, even if appropriate exposure values are given,
  it is difficult to set them at the time of photographing.
  In particular, if the scene is dynamic or the camera moves while pictures
  are being captured, the exposure time should be shortened
  to prevent ghost-like or blurring artifacts in the fused image.
  The literature \cite{kinoshita2018multi} has pointed out that
  it is possible to improve the quality of multi-exposure images
  by adjusting the luminance of the images after photographing,
  but how to determine the degree has never been discussed.
  
  To overcome these problems, this paper proposes a novel luminance adjustment method
  based on automatic exposure compensation for multi-exposure image fusion.
  The proposed method automatically determines the degree of adjustment
  on the basis of the luminance distribution of input multi-exposure images.
  Moreover, the proposed luminance adjustment method enables us to produce
  high quality images from the adjusted ones by a fusion method with simple weights,
  although the adjusted ones can be combined by any existing fusion methods.
  
  We evaluate the effectiveness of the proposed method by using various fusion methods.
  Experimental results show that the multi-exposure images adjusted by the proposed method
  have better quality than the input multi-exposure ones in terms of the well-exposedness.
  The results also denote that the proposed method enables
  to produce fused images with high quality under various fusion methods,
  and moreover, the proposed simple weights provide the highest score of
  statistical naturalness and discrete entropy in all fusion methods.
\section{Preparation}
  Existing multi-exposure fusion methods use images taken under different
  exposure conditions, i.e., ``multi-exposure images.''
  Here we discuss the relationship between exposure values and pixel values.
  For simplicity, we focus on grayscale images in this section.
\subsection{Relationship between exposure values and pixel values}
\begin{figure}[!t]
    \centering
    \includegraphics[width=0.95\linewidth]{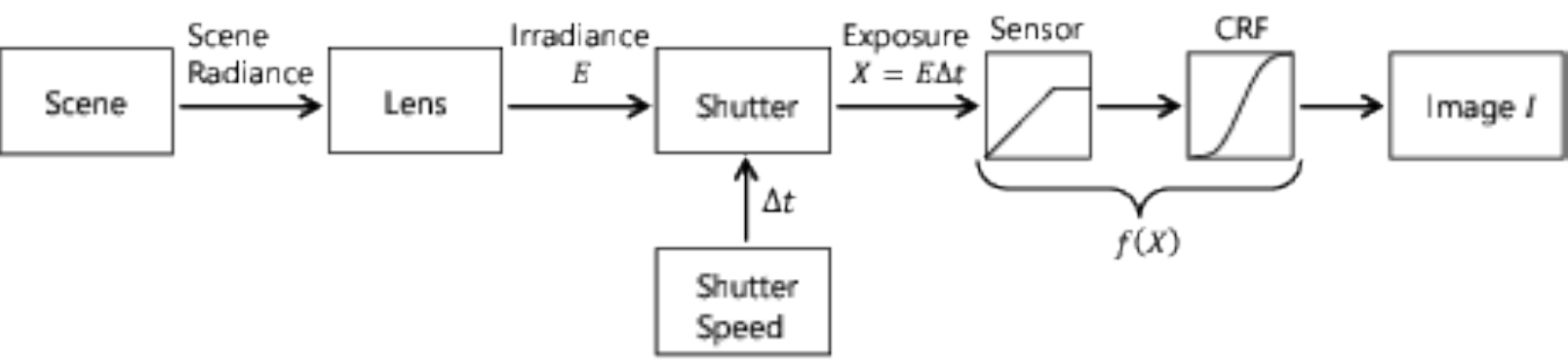}
    \caption{Imaging pipeline of digital camera \label{fig:camera}}
  \end{figure}

  Figure \ref{fig:camera} shows a typical imaging pipeline for
  a digital camera\cite{dufaux2016high}.
  The radiant power density at the sensor, i.e., irradiance $E$,
  is integrated over the time $\Delta t$ the shutter is open,
  producing an energy density, commonly referred to as exposure $X$.
  If the scene is static during this integration,
  exposure $X$ can be written simply as the product of irradiance $E$ and
  integration time $\Delta t$ (referred to as ``shutter speed''):
  \begin{equation}
    X(p) = E(p)\Delta t,
    \label{eq:exposure}
  \end{equation}
  where $p=(x,y)$ indicates the pixel at point $(x,y)$.
  A pixel value $I(p) \in [0, 1]$ in the output image $I$ is given by
  \begin{equation}
    I(p) = f(X(p)),
    \label{eq:CRF}
  \end{equation}
  where $f$ is a function combining sensor saturation
  and a camera response function (CRF).
  The CRF represents the processing in each camera which makes the final image $I(p)$
  look better.

  Camera parameters, such as shutter speed and lens aperture,
  are usually calibrated in terms of exposure value (EV) units,
  and the proper exposure for a scene is automatically decided by
  the camera.
  The exposure value is commonly controlled by changing the shutter speed
  although it can also be controlled by adjusting various camera parameters.
  Here we assume that the camera parameters except for
  the shutter speed are fixed.
  Let $\nu = 0 \mathrm{[EV]}$ and $\Delta \tau$
  be the proper exposure value
  and shutter speed under the given conditions, respectively.
  The exposure value $v_i \mathrm{[EV]}$ of an image taken at
  shutter speed $\Delta t_i$ is given by
  \begin{equation}
    v_i = \log_2 \Delta t_i - \log_2 \Delta \tau.
    \label{eq:EV}
  \end{equation}
  From (\ref{eq:exposure}) to (\ref{eq:EV}),
  images $I_0$ and $I_i$ exposed at $0 \mathrm{[EV]}$ and $v_i \mathrm{[EV]}$,
  respectively, are written as
  \begin{align}
    I_0(p) &= f(E(p)\Delta \tau) \label{eq:CRFwithExposure}\\
    I_i(p) &= f(E(p)\Delta t_i) \label{eq:CRFwithExposure2}
            = f(2^{v_i} E(p)\Delta \tau).
  \end{align}
  Assuming function $f$ is linear,
  we obtain the following relationship between $I_0$ and $I_i$:
  \begin{equation}
    I_i(p) = 2^{v_i} I_0(p).
    \label{eq:relationship}
  \end{equation}
  Therefore, the exposure can be varied artificially by multiplying $I_0$ by a constant.
  This ability is used in a new multi-exposure fusion method,
  which is described in the next section.

\subsection{Scenario}
  For multi-exposure fusion methods to produce high quality images,
  the input images should represent the bright, middle, and dark
  regions of the scene.
  These images generally consist of
  a properly exposed image ($v_i = 0 \mathrm{[EV]}$),
  overexposed images ($v_i > 0$), and underexposed images ($v_i < 0$).
  For example, three multi-exposure images might be taken at $v_i = -1, 0, +1 \mathrm{[EV]}$.
  However, there are several problems in photographing multi-exposure images as follows:
  \begin{itemize}[nosep]
    \item Determining appropriate exposure values
      for multi-exposure image fusion.
    \item Setting appropriate exposure values under the time of photographing
      when there are time constraints.
    \item Using an image taken at $0 \mathrm{[EV]}$ that
      might not represent the scene properly.
  \end{itemize}

\begin{figure}[!t]
  \centering
  \subfloat[Input images $I_f$ ``Window'' ($v_i=-1, 0, +1\mathrm{[EV]}$)]{
    \includegraphics[width=0.30\hsize]{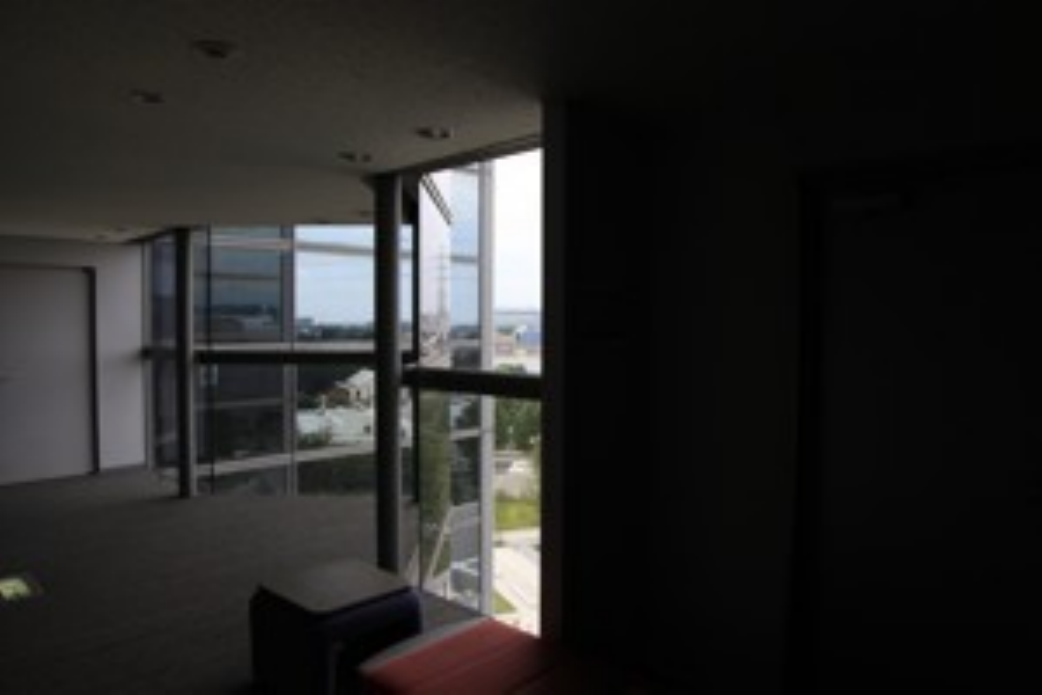}
    \includegraphics[width=0.30\hsize]{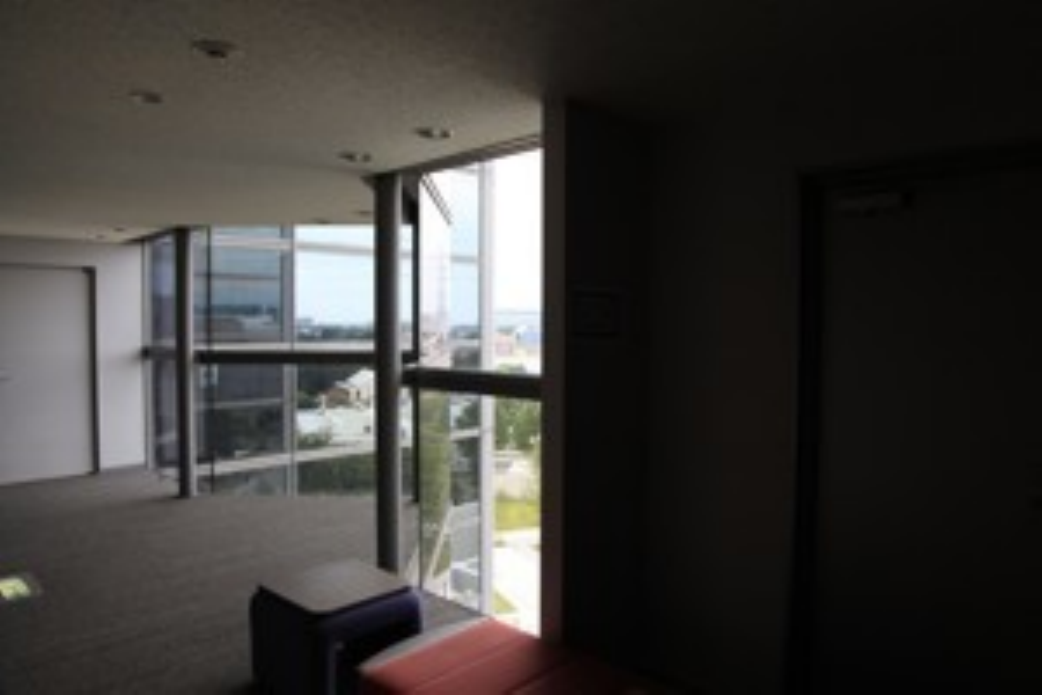}
    \includegraphics[width=0.30\hsize]{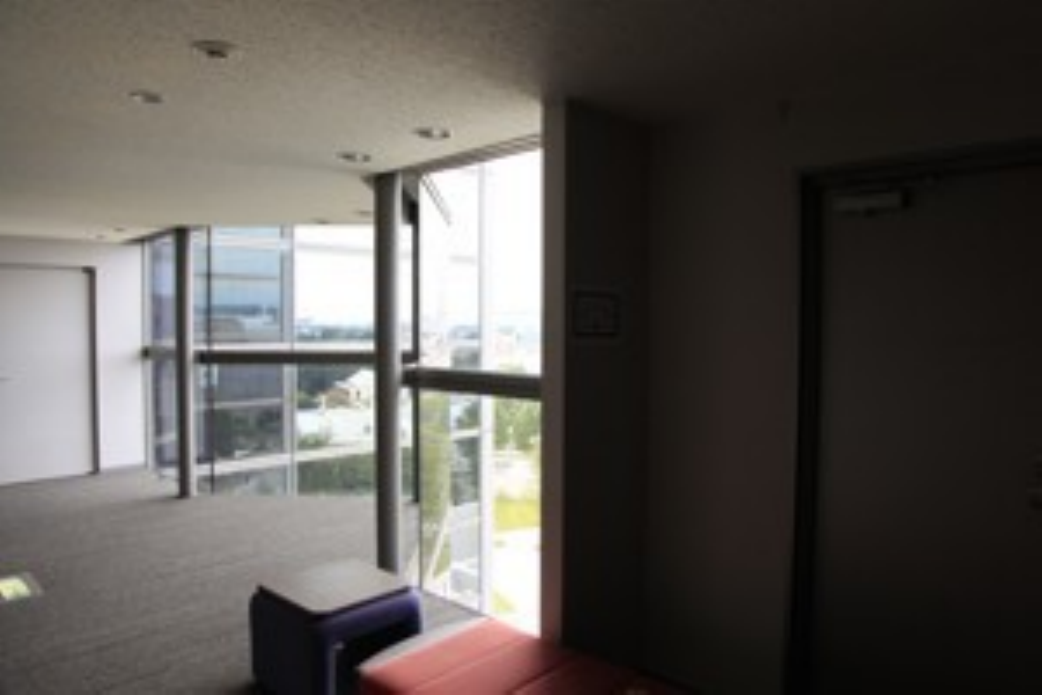}
    \label{fig:window_input}}\\
  \subfloat[Adjusted example images 1]{
    \includegraphics[width=0.30\hsize]{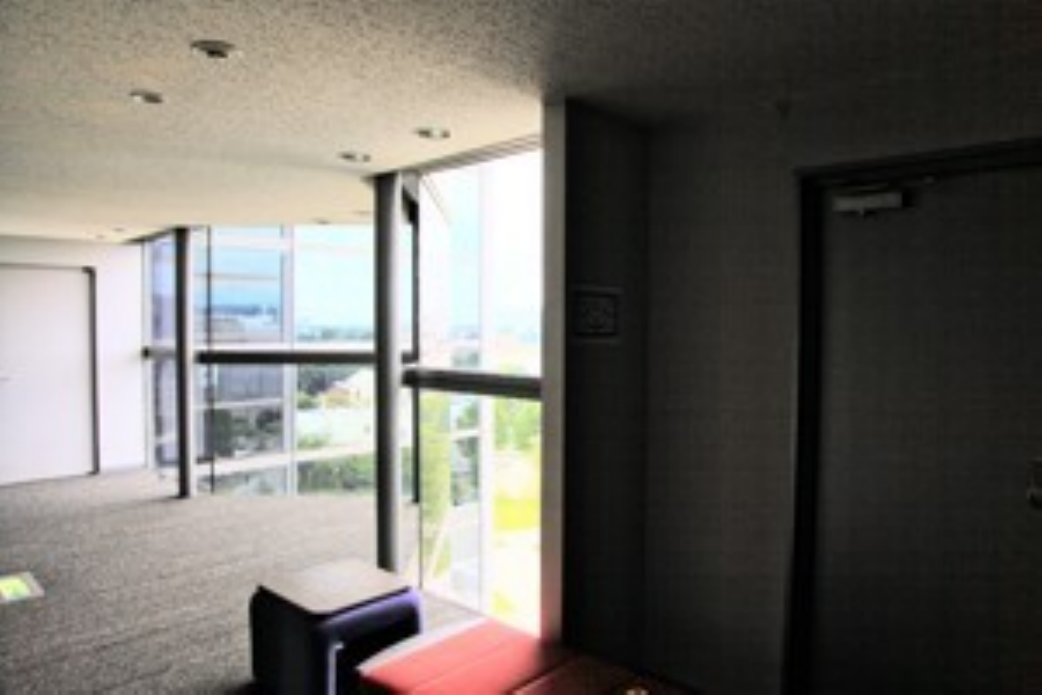}
    \includegraphics[width=0.30\hsize]{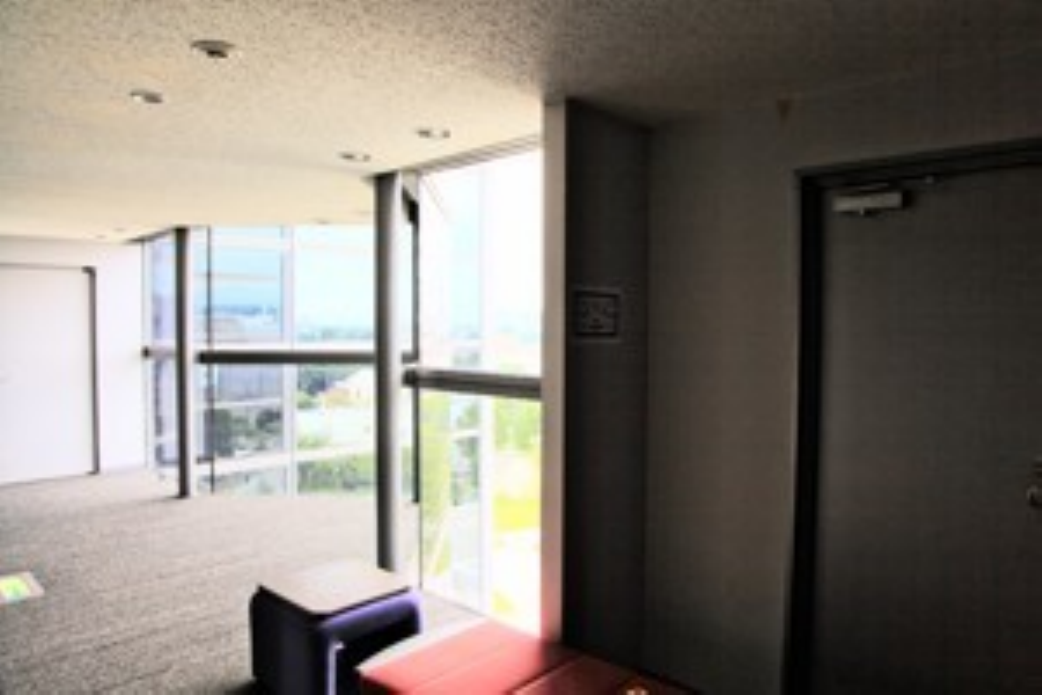}
    \includegraphics[width=0.30\hsize]{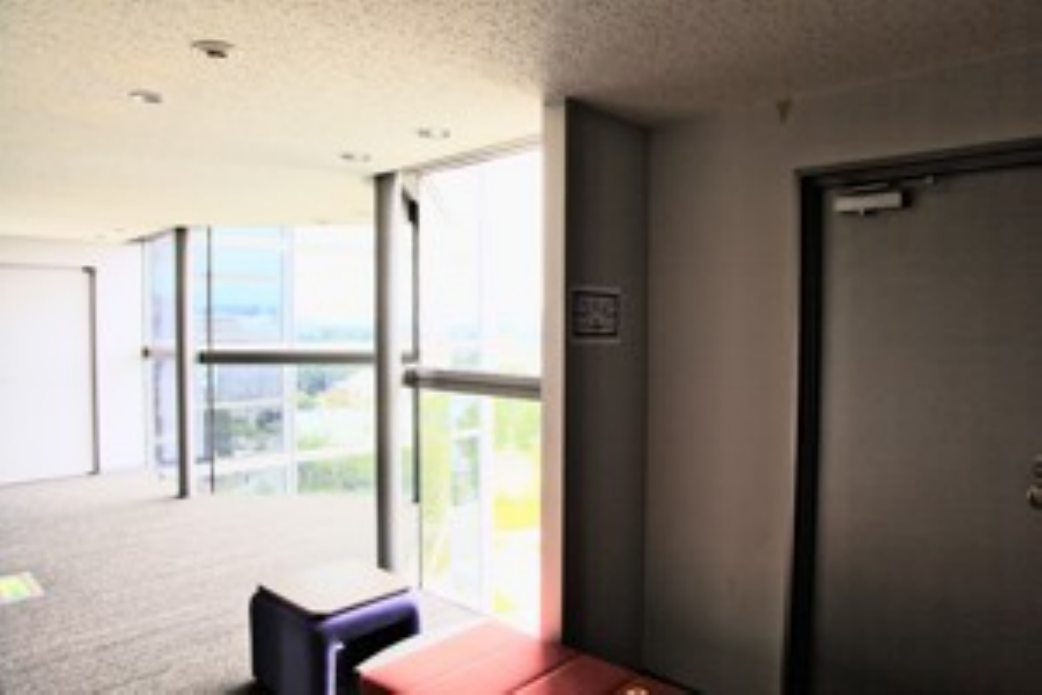}
    \label{fig:window_enhanced_conv}}\\
  \subfloat[Adjusted example images 2]{
    \includegraphics[width=0.30\hsize]{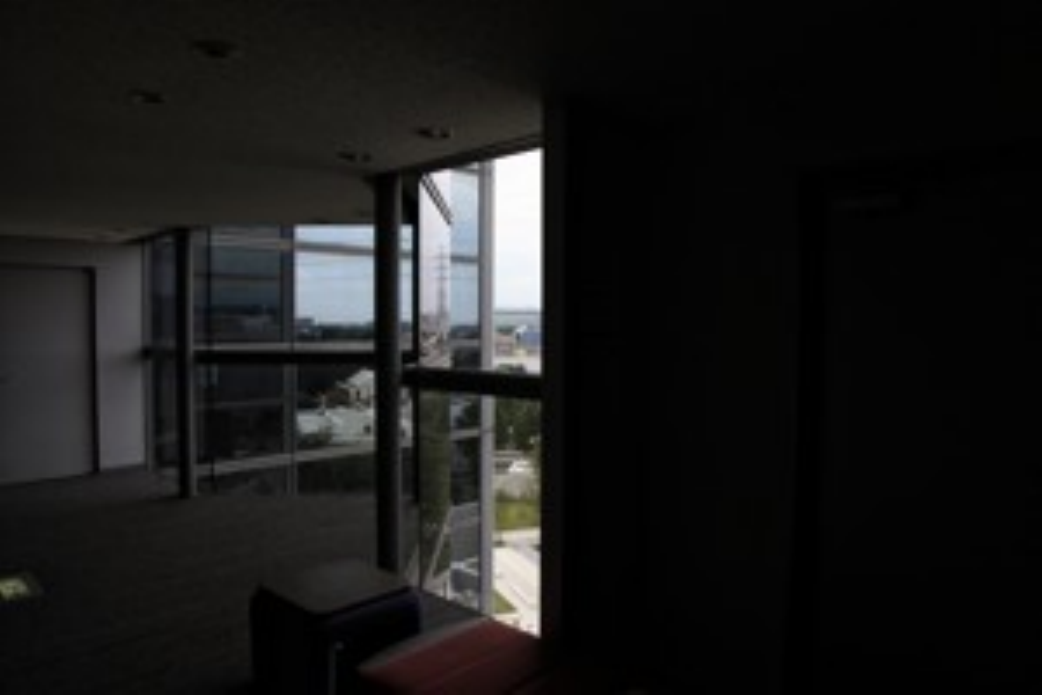}
    \includegraphics[width=0.30\hsize]{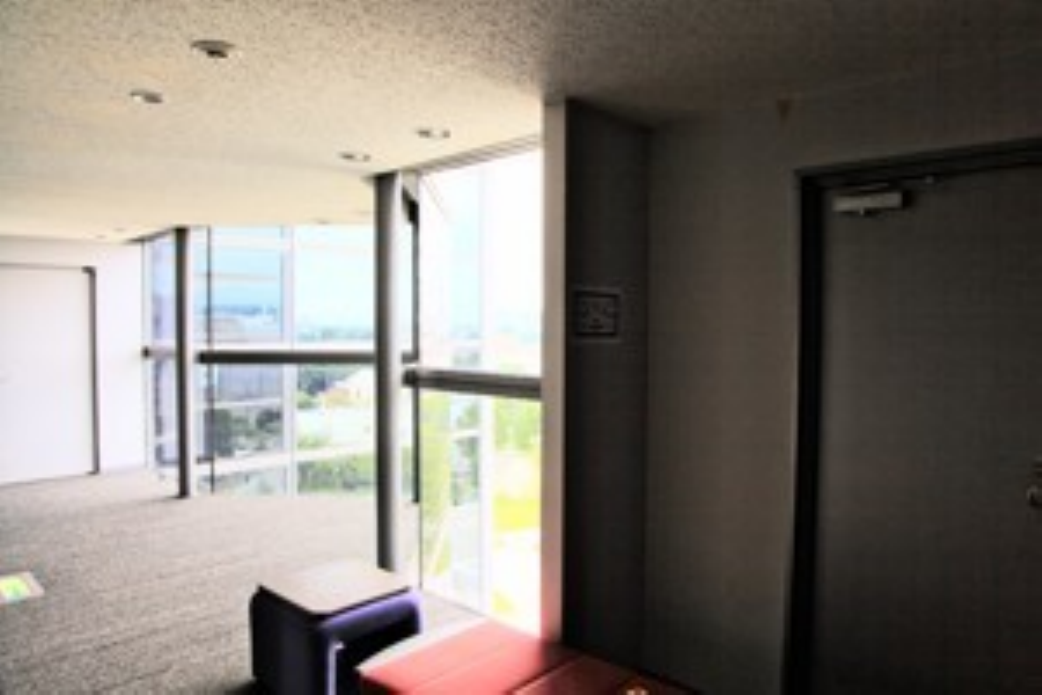}
    \includegraphics[width=0.30\hsize]{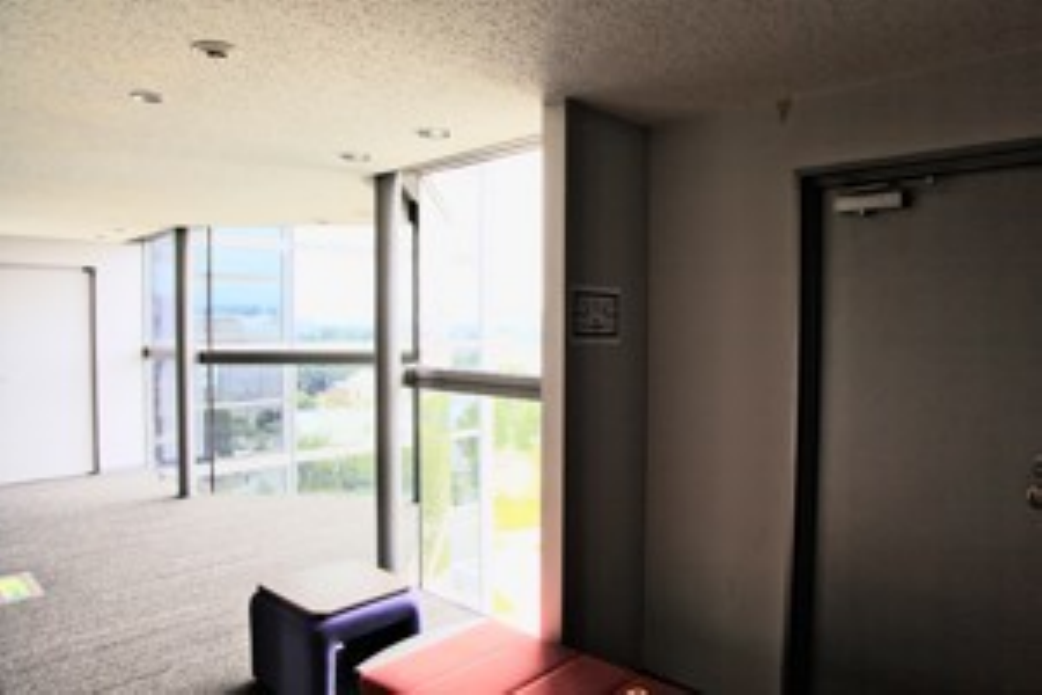}
    \label{fig:window_enhanced_prop}}\\
  \caption{Examples of adjusted multi-exposure images}
  \label{fig:inputImages}
\end{figure}

\begin{figure}[!t]
  \centering
  \subfloat[Input]{
    \includegraphics[width=0.30\hsize]{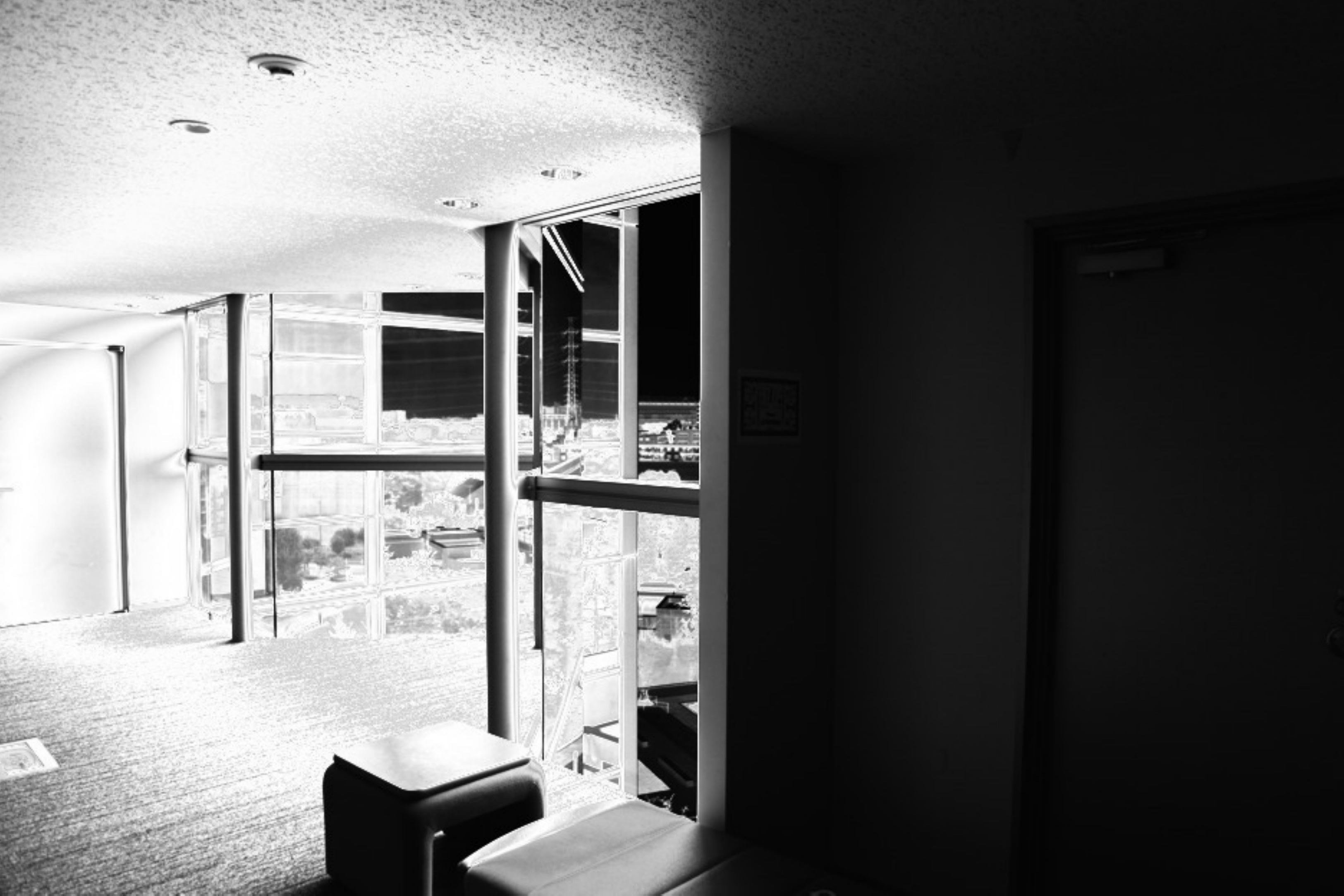}
    \label{fig:well_exposedness_orig}
  }
  \subfloat[Fig. 2(b)]{
    \includegraphics[width=0.30\hsize]{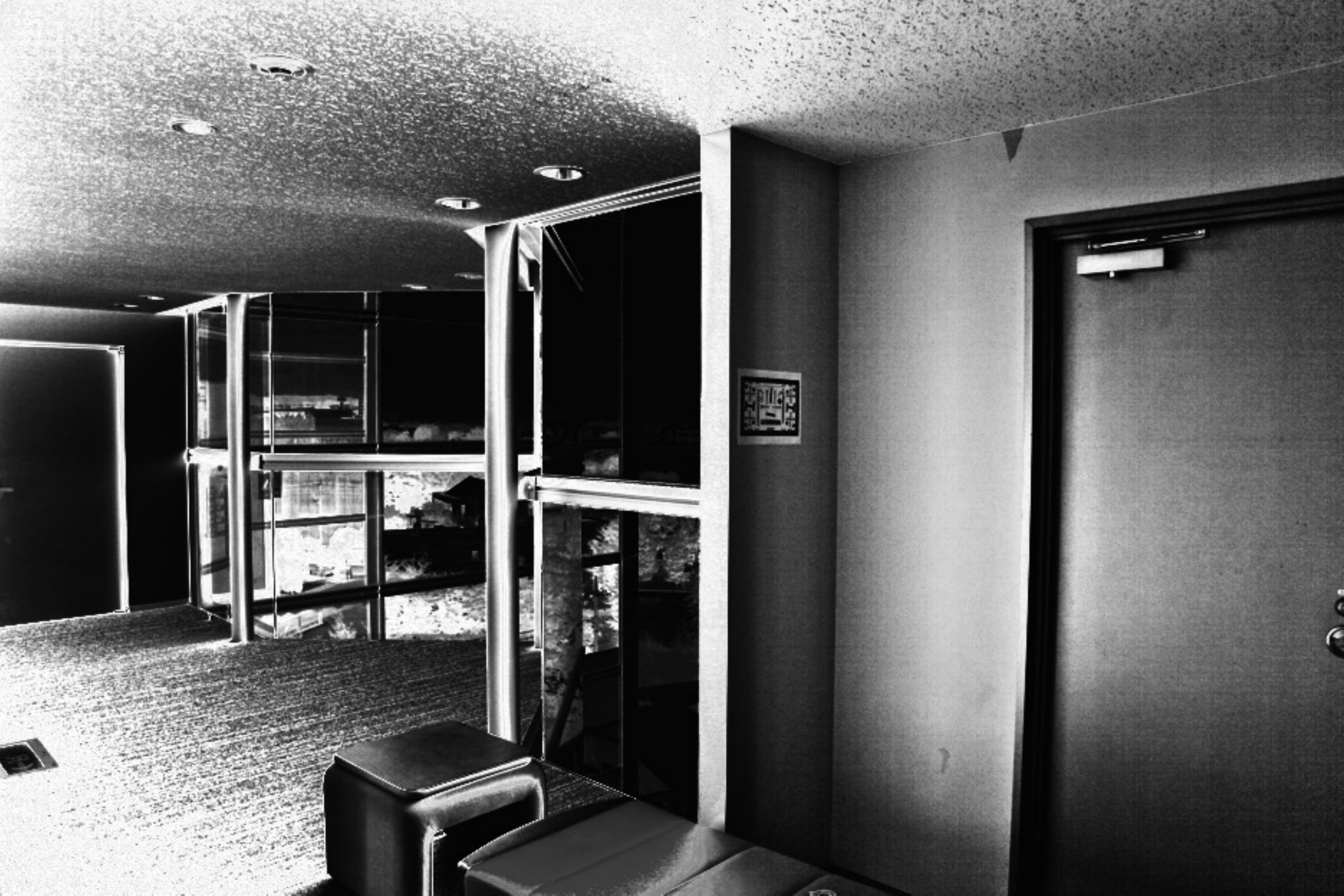}
    \label{fig:well_exposedness_conv}
  }
  \subfloat[Fig. 2(c)]{
    \includegraphics[width=0.30\hsize]{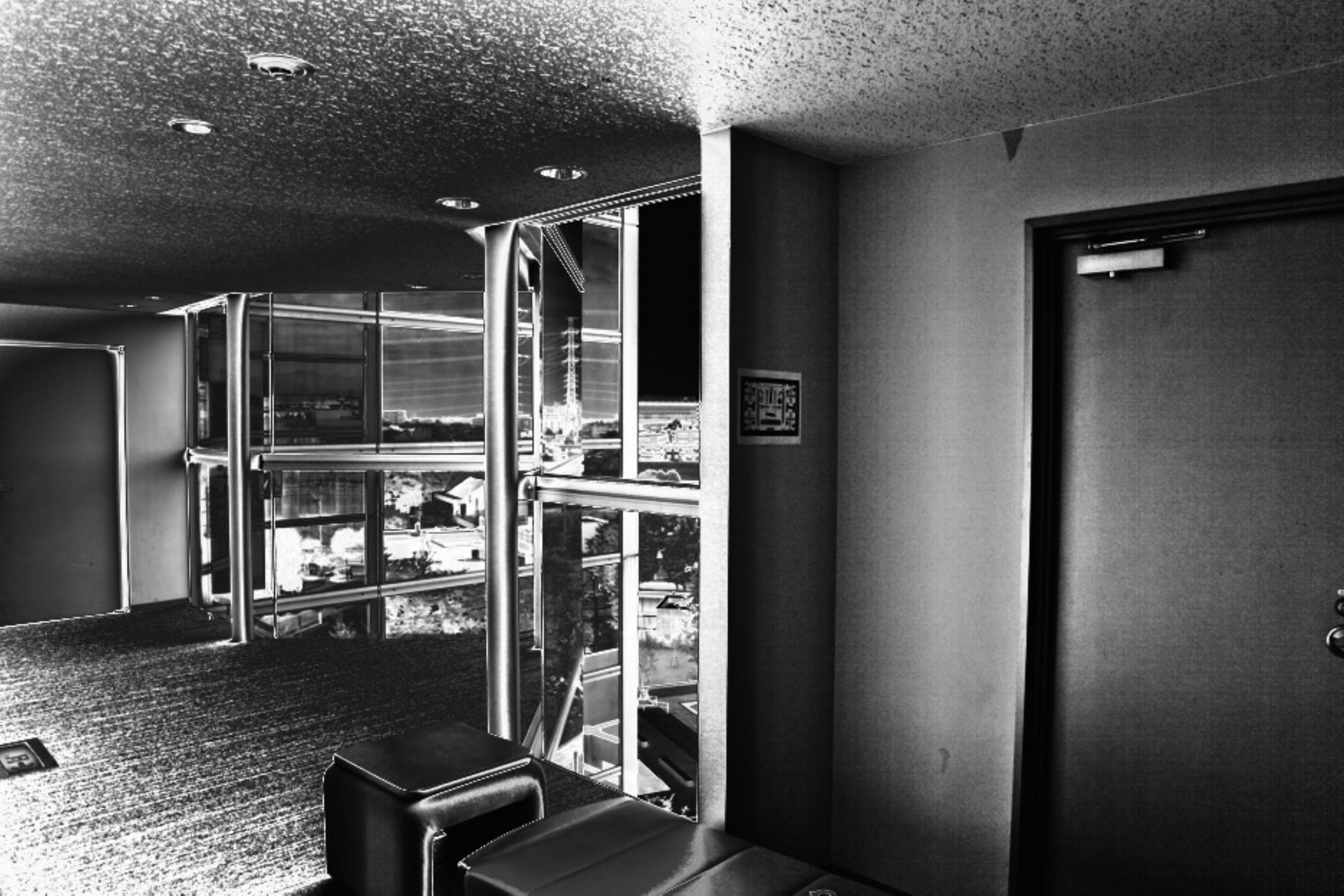}
    \label{fig:well_exposedness_prop}
  }\\
  \caption{Well-exposedness map for input and enhanced multi-exposure images.
  A brighter pixel indicates that the pixel is well exposed.}
  \label{fig:well_exposedness}
\end{figure}

  The literature \cite{kinoshita2018multi} pointed out that
  it is possible to improve the quality of multi-exposure images
  by adjusting the luminance of the images.
  Figure \ref{fig:inputImages} shows examples of adjusted multi-exposure ones.
  The quality of multi-exposure images can be evaluated by using the well-exposedness
  \cite{mertens2009exposure},
  which indicates how well a pixel is exposed.
  Figure \ref{fig:well_exposedness} shows the maximum score of the well-exposedness
  for each pixel in multi-exposure images.
  These results in Figs. \ref{fig:inputImages} and \ref{fig:well_exposedness} denote
  that the quality of multi-exposure images depends on the degree of adjustment.
  However, how to determine the degree has never been discussed.

  Because of such a situation, this paper proposes a new luminance adjustment method based on
  automatic exposure compensation for multi-exposure fusion.
  In addition, we look at appropriate multi-exposure fusion methods
  for the adjusted multi-exposure images.

\section{Proposed luminance adjustment method}
\begin{figure*}[!t]
    \centering
    \includegraphics[clip, width=12cm]{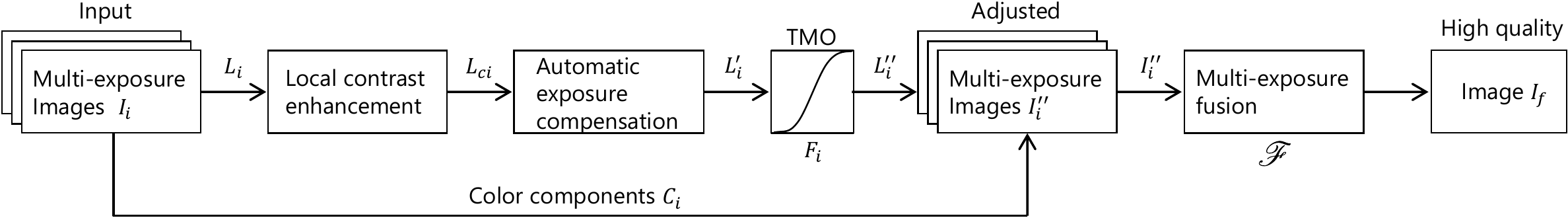}
    \caption{Use of proposed luminance adjustment method
      in multi-exposure image fusion \label{fig:proposedMEF}}
  \end{figure*}

  The use of the proposed method in multi-exposure fusion is illustrated
  in Fig. \ref{fig:proposedMEF}.
  To enhance the quality of multi-exposure images, local contrast enhancement is applied to
  luminance $L_i (1 \le i \le N, i \in \mathbb{N})$
  calculated from the $i$-th input image $I_i$, and then
  automatic exposure compensation and tone mapping are applied.
  Next, image $I_f$ with improved quality is produced by multi-exposure image fusion methods
  such as a weighted average.
  Here we consider input image $I_i$ with exposure value $v_i$
  that satisfies $v_i < v_{i+1}$.
\subsection{Local contrast enhancement}
  If the input images do not represent the scene clearly,
  the quality of an image fused from them will be lower than
  that of an image fused from ideally exposed images.
  Therefore, the dodging and burning algorithm is used to enhance
  the local contrast\cite{huo2013dodging}.
  The luminance $L_{ci}$ enhanced by the algorithm is given by
  \begin{equation}
    L_{ci}(p) = \frac{L_i^2(p)}{L_{ai}(p)},
    \label{eq:dodgingAndBurning}
  \end{equation}
  where $L_{ai}(p)$ is the local average of luminance $L_i(p)$ around pixel $p$.
  It is obtained by applying a low-pass filter to $L_i(p)$.
  Here, a bilateral filter is used for this purpose.

  $L_{ai}(p)$ is calculated using the bilateral filter:
  \begin{equation}
    L_{ai}(p) = \frac{1}{c_i(p)}
              \sum_{q \in \Omega}
                L_i(q) g_{\sigma_1}(q-p) g_{\sigma_2}(L_i(q) - L_i(p)),
    \label{eq:bilateral}
  \end{equation}
  where $\Omega$ is the set of all pixels, and $c_i(p)$ is a normalization term such as
  \begin{equation}
    c_i(p) = \sum_{q \in \Omega} g_{\sigma_1}(q-p) g_{\sigma_2}(L_i(q) - L_i(p)),
    \label{eq:normalizingConst}
  \end{equation}
  where $g_{\sigma}$ is a Gaussian function given by
  \begin{equation}
    g_{\sigma}(p | p=(x,y)) = C_{\sigma}\exp \left( -\frac{x^2 + y^2}{\sigma^2} \right)
    \label{eq:gaussian}
  \end{equation}
  using a normalization factor $C_{\sigma}$.
  Parameters $\sigma_1 = 16$ and $\sigma_2 = 3/255$ are set
  in accordance with \cite{huo2013dodging}.
\subsection{Automatic exposure compensation}
  The purpose of the proposed exposure compensation is to automatically adjust the luminance of
  each input image $I_i$, so that adjusted images have appropriate exposure values
  for multi-exposure image fusion.
  The luminance $L'_i$ of adjusted image $I'_i$ is simply obtained by,
  according to eq. (\ref{eq:relationship}),
  \begin{equation}
    L'_i(p) = \alpha_i L_{ci}(p),
    \label{eq:constMultiplication}
  \end{equation}
  where parameter $\alpha_i > 0$ indicates the degree of adjustment.
  Next, the way to estimate the parameter $\alpha_i$ is described.

  In $N$ input images, the $j = \lceil \frac{N + 1}{2} \rceil$-th image $I_j$ has middle
  brightness, and the overexposed (or underexposed) areas in $I_j$ are smaller than
  those in the other images. Therefore, the quality of image $I_j$
  should be better than that of the other images.
  We thus estimate parameter $\alpha_j$ from the $j$-th image in order to
  map the geometric mean $\overline{L}_{cj}$ of luminance $L_{cj}$ to middle-gray
  of the displayed image, or 0.18 on a scale from zero to one,
  as in \cite{reinhard2002photographic},
  where the geometric mean of the luminance values indicates
  the approximate brightness of the image.
  
  Let $P$ and $L(p)$ are a subset of $\Omega$ and the luminance of $p \in P$,
  respectively.
  Then the geometric mean $G(L|P)$ of luminance $L(p)$ is calculated using
  \begin{equation}
    G(L|P) =
      \exp{
        \left(\frac{1}{|P|}
          \sum_{p \in P} \log{\left(\max{\left( L(p), \epsilon \right)}\right)}
        \right)
      },
    \label{eq:geoMeanEps}
  \end{equation}
  where $\epsilon$ is set to a small value to avoid singularities at $L(p)=0$.
  Parameter $\alpha_j$ is derived using eq. (\ref{eq:geoMeanEps}) from
  \begin{equation}
    \alpha_j = \frac{0.18}{G(L_{cj}|\Omega)}.
    \label{eq:alphaj}
  \end{equation}

  The adjusted version $I'_k$ of the $k$-th input image $I_k (k \neq j)$ should describes
  some areas that $I'_j$ could not represent well.
  Such areas are overexposed and underexposed regions in $I_j$.
  For this reason, we divide the luminance range of $I_j$ into $N$ equal parts
  $P_1, \cdots, P_N$ as
  \begin{equation}
    P_k = \{p|\theta_k \le L_{cj}(p) \le \theta_{k+1}\},
    \label{eq:luminance_part}
  \end{equation}
  where $\theta_k$ is calculated as
  \begin{equation}
    \theta_k = \frac{N-k+1}{N}\left(\max L_{cj}(p) - \min L_{cj}(p)\right) + \min L_{cj}(p).
  \end{equation}
  Note that $P_k$ satisfies $\Omega = P_1 \cup P_2 \cup \cdots \cup P_N$.
  Then we adjust $I_k$ so that it could represent the $k$-th brightest part $P_k$ well.
  By using eqs. (\ref{eq:geoMeanEps}) and (\ref{eq:luminance_part}),
  parameter $\alpha_k$ is calculated as
  \begin{equation}
    \alpha_k = \frac{0.18}{G(L_{ck}|P_k)}.
    \label{eq:unknownEV}
  \end{equation}
  Eq. (\ref{eq:unknownEV}) enables us to produce multi-exposure images
  that represent not only dark areas but also bright areas.
\subsection{Tone mapping}
  Since the adjusted luminance value $L'_i(p)$ often exceeds
  the maximum value of the common image format,
  pixel values might be lost due to truncation of the values.
  This problem is overcome by using a tone mapping operation
  to fit the adjusted luminance value into the interval $[0, 1]$.

  The luminance $L''_i$ of an enhanced multi-exposure image is obtained
  by applying a tone mapping operator $F_i$ to $L'_i$:
  \begin{equation}
    L''_i(p) = F_i(L'_i(p)).
    \label{eq:TM}
  \end{equation}
  Reinhard's global operator is used here as a tone mapping operator $F_i$
  \cite{reinhard2002photographic}.
  
  Reinhard's global operator is given by
  \begin{equation}
    F_i(L(p)) = \frac{L(p)}{1 + L(p)}\left(1 + \frac{L(p)}{L^2_{white_i}} \right),
    \label{eq:reinhardTMO}
  \end{equation}
  where parameter $L_{white_i} > 0$ determines luminance value $L(p)$
  as $F_i(L(p)) = 1$.
  Note that Reinhard's global operator $F_i$ is a monotonically increasing function.
  Here, let $L_{white_i} = \max L'_i(p)$.
  We obtain $L''_i(p) \le 1$ for all $p$.
  Therefore, truncation of the luminance values can be prevented.

  Combining $L''_i$,
  luminance $L_i$ of the $i$-th input image $I_i$,
  and RGB pixel values $C_i(p) \in \{R_i(p), G_i(p), B_i(p)\}$ of $I_i$,
  we obtain RGB pixel values $C''_i(p) \in \{R''_i(p), G''_i(p), B''_i(p)\}$ of
  the enhanced multi-exposure images $I''_i$:
  \begin{equation}
    C''_i(p) = \frac{L''_i(p)}{L_i(p)}C_i(p).
    \label{eq:color}
  \end{equation}
\subsection{Fusion of enhanced multi-exposure images}
  Enhanced multi-exposure images $I''_i$ can be used as input for
  any existing multi-exposure image fusion methods.
  A final image $I_f$ is produced as
  \begin{equation}
    I_f(p) = \mathscr{F}(I''_1(p), I''_2(p), \cdots, I''_N(p)),
    \label{eq:fusion}
  \end{equation}
  where $\mathscr{F}(I_1(p), I_2(p), \cdots, I_N(p))$ indicates a function to fuse $N$ images
  $I_1, I_2, \cdots, I_N$ into a single image.

  While numerous methods $\mathscr{F}$ for fusing images have been proposed,
  methods based on a weighted average are widely used
  \cite{mertens2009exposure,nejati2017fast}
  and the weighted average is calculated as
  \begin{equation}
    \mathscr{F}(I_1(p), I_2(p), \cdots, I_N(p))
    = \frac{\sum_{i=1}^N w_i(p) I_i(p)}{\sum_{i=1}^N w_i(p)}.
    \label{eq:weighted_average}
  \end{equation}
  Eq. (\ref{eq:weighted_average}) aims to produce high quality images
  by adjusting weights $w_i(p)$
  under the condition that pixel values $I_i(p)$ are fixed.
  For example, the weight $w_i(p)$ is calculated on the basis of contrast, color saturation,
  and well-exposedness of each pixel, as in \cite{mertens2009exposure}.
  On the other hand, in the proposed method, pixel values $I_i(p)$ are adjusted by
  considering well-exposedness before the fusion.
  Therefore, the proposed method enables us to use simpler weights,
  like $w_i(p) = 1$ referred to as
  ``simple average'', although conventional weights are also available.
  In the next section, it will be shown that the simple average provides better results
  than conventional weights for the proposed luminance adjustment method.
\section{Simulation}
  We evaluated the effectiveness of the proposed luminance adjustment method
  in terms of the quality of generated images $I_f$ and adjusted multi-exposure images $I''_i$.
\subsection{Comparison with conventional methods}
  To evaluate the quality of the images produced by each method,
  objective quality assessments are needed.
  Typical quality assessments such as the peak signal to noise ratio (PSNR)
  and the structural similarity index (SSIM) are not suitable for this purpose
  because they use the target image with the highest quality as the reference one.
  We therefore used the tone mapped image quality index (TMQI) \cite{yeganeh2013objective}
  and discrete entropy as quality assessments.
  In addition, we utilized the well-exposedness to measure the quality
  of adjusted multi-exposure images, as in 2.2.

  TMQI represents the quality of an image tone mapped from an HDR image;
  the index incorporates structural fidelity and statistical naturalness.
  An HDR image is used as a reference to calculate structural fidelity.
  A reference is not needed to calculate statistical naturalness.
  Since the processes of tone mapping and photographing are similar,
  TMQI is also useful for evaluating photographs.
  Discrete entropy represents the amount of information in an image.
\subsection{Simulation conditions}
  In the simulation,
  four photographs taken by Canon EOS 5D Mark II camera
  and eight photographs selected from an available online database \cite{easyhdr}
  were used as input images $I_i$
  (see Fig. \ref{fig:inputImages}\subref{fig:window_input}).
  The following procedure was carried out to evaluate the effectiveness.
  \begin{enumerate}[nosep]
    \item Produce $I''_i$ from $I_i$ using the proposed method.
    \item Obtain $I_f$ fused from $I''_i$ by $\mathscr{F}$.
    \item Compute the well-exposedness of $I''_i$.
    \item Compute TMQI values between $I_f$ and $I_H$.
    \item Compute discrete entropy of $I_f$.
  \end{enumerate}
  Here we used four fusion methods, i.e., Mertens' method \cite{mertens2009exposure},
  Sakai's method \cite{sakai2015hybrid}, Nejati's method \cite{nejati2017fast}, and
  the simple average, as $\mathscr{F}$.

  In addition, structural fidelity in the TMQI could not be calculated due to
  the non-use of HDR images.
  Thus, we used only statistical naturalness in the TMQI for the evaluation.
\subsection{Simulation results}
\begin{figure}[!t]
  \centering
  \subfloat{
    \includegraphics[width=0.30\hsize]{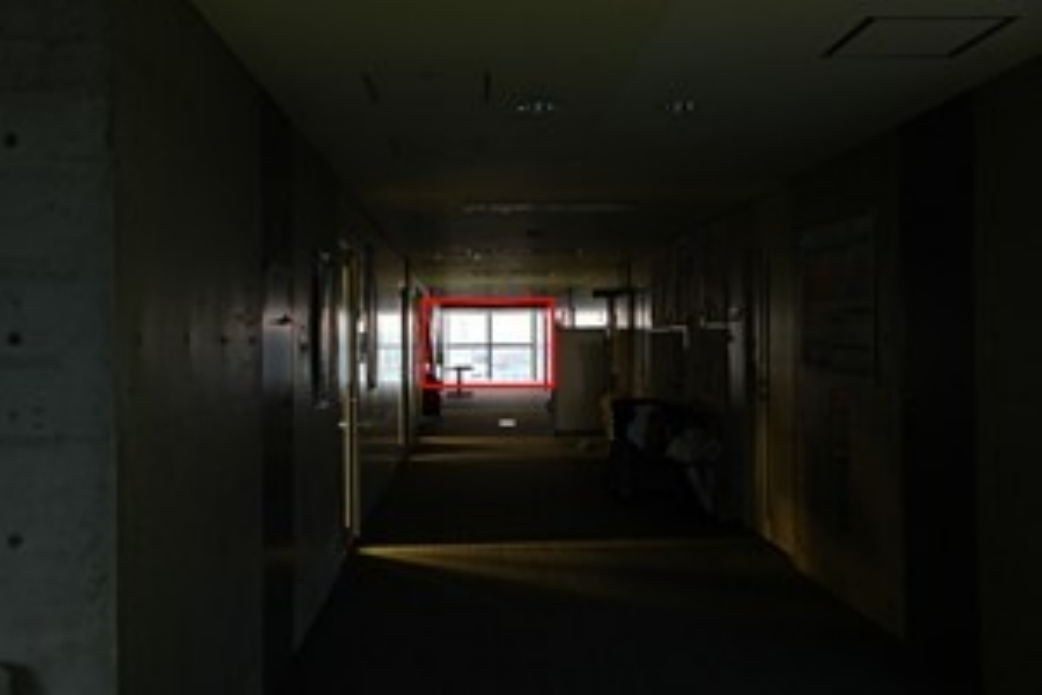}
  }
  \subfloat{
    \includegraphics[width=0.30\hsize]{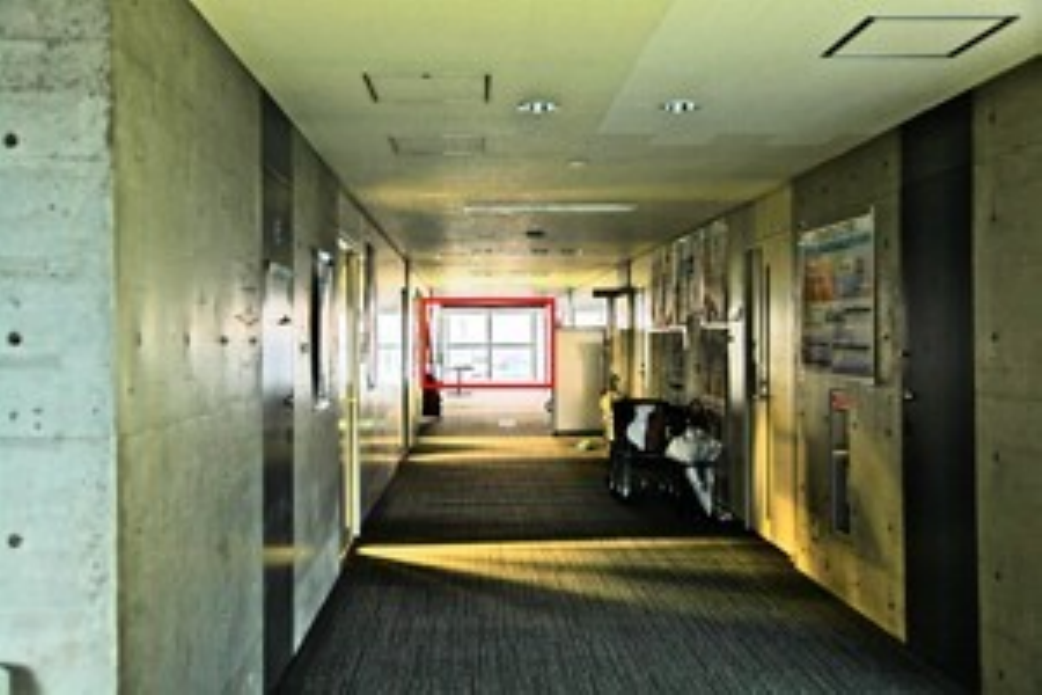}
  }\\
  \addtocounter{subfigure}{-2}
  \subfloat{
    \includegraphics[width=0.30\hsize]{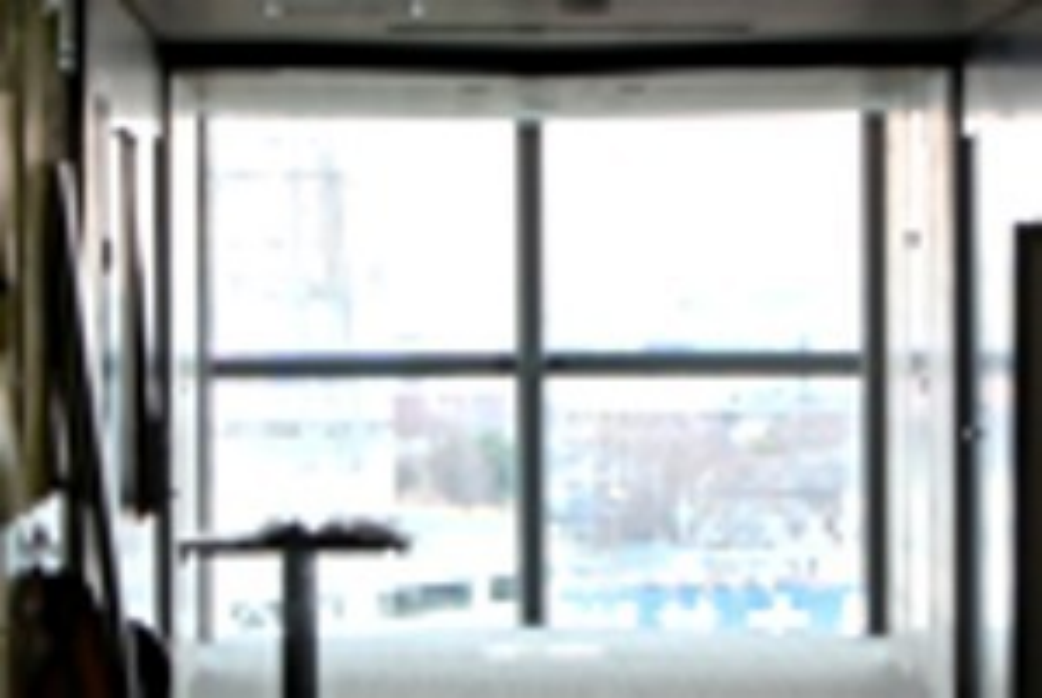}
  }
  \subfloat{
    \includegraphics[width=0.30\hsize]{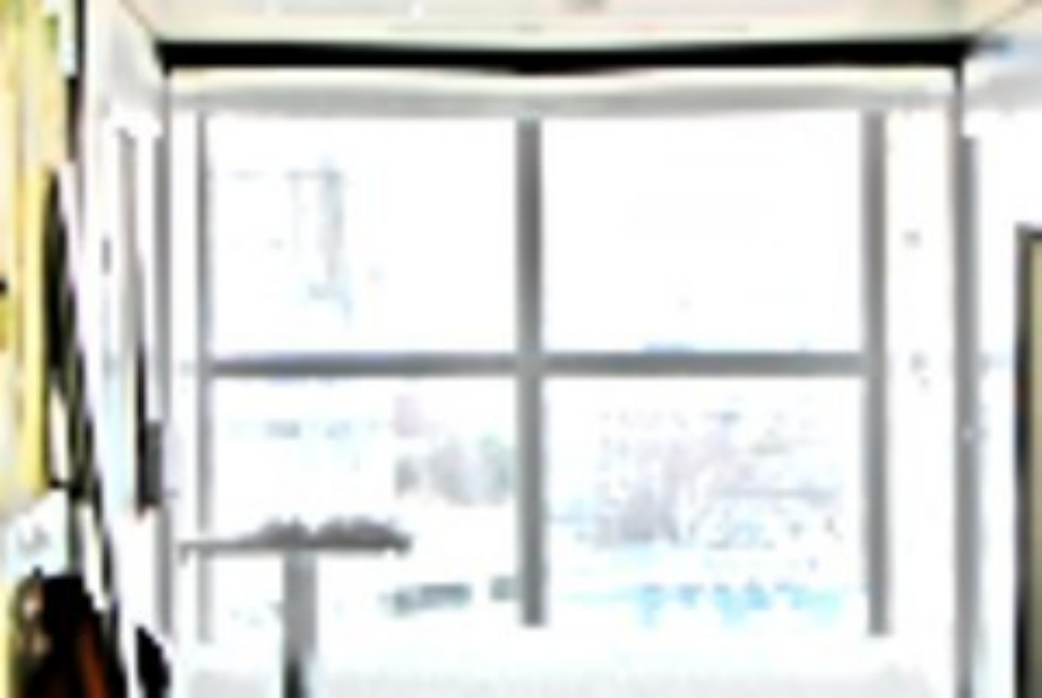}
  }\\
  \addtocounter{subfigure}{-2}
  \subfloat[Without adjustment]{
    \includegraphics[width=0.30\hsize]{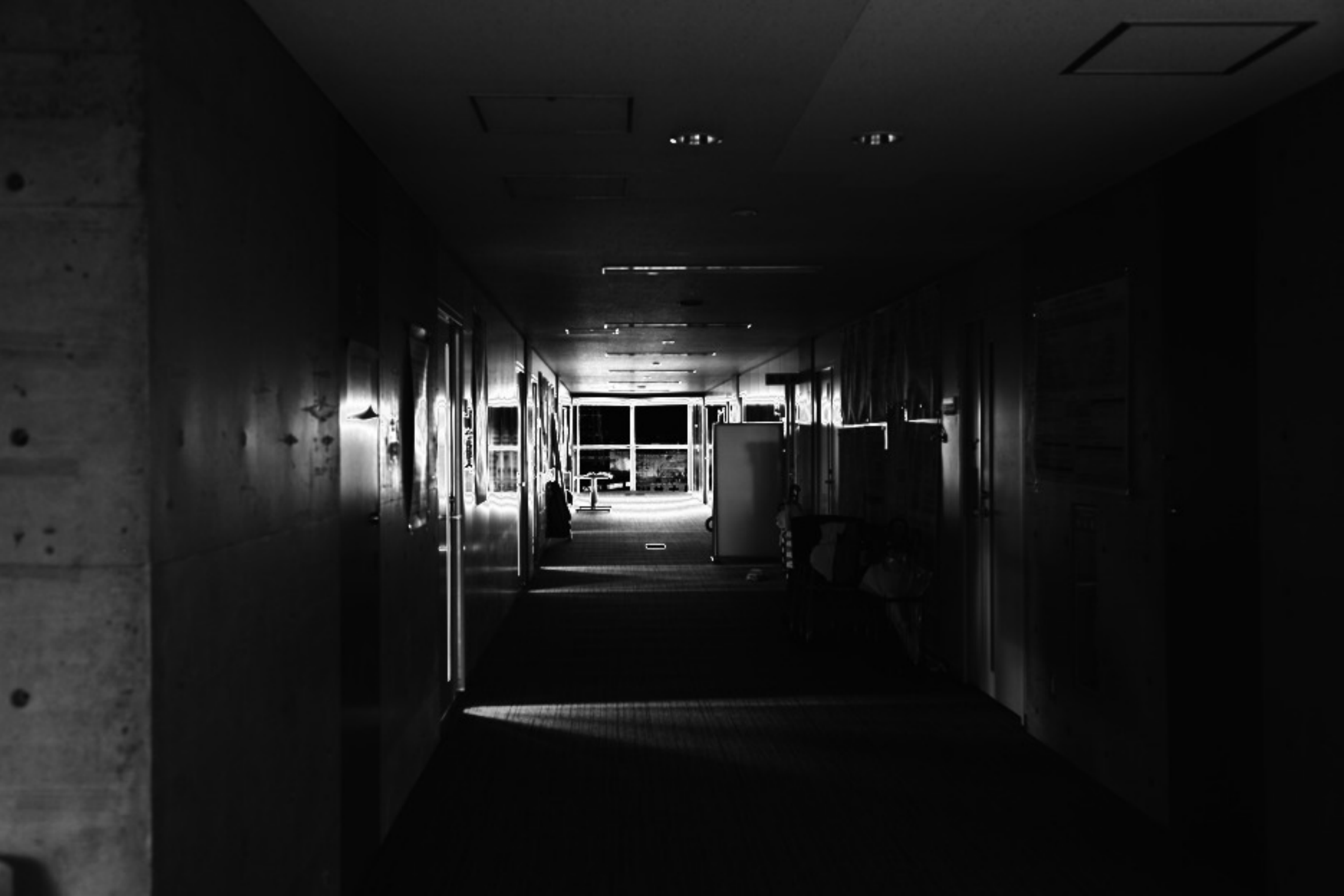}
  }
  \subfloat[Adjusted by ours]{
    \includegraphics[width=0.30\hsize]{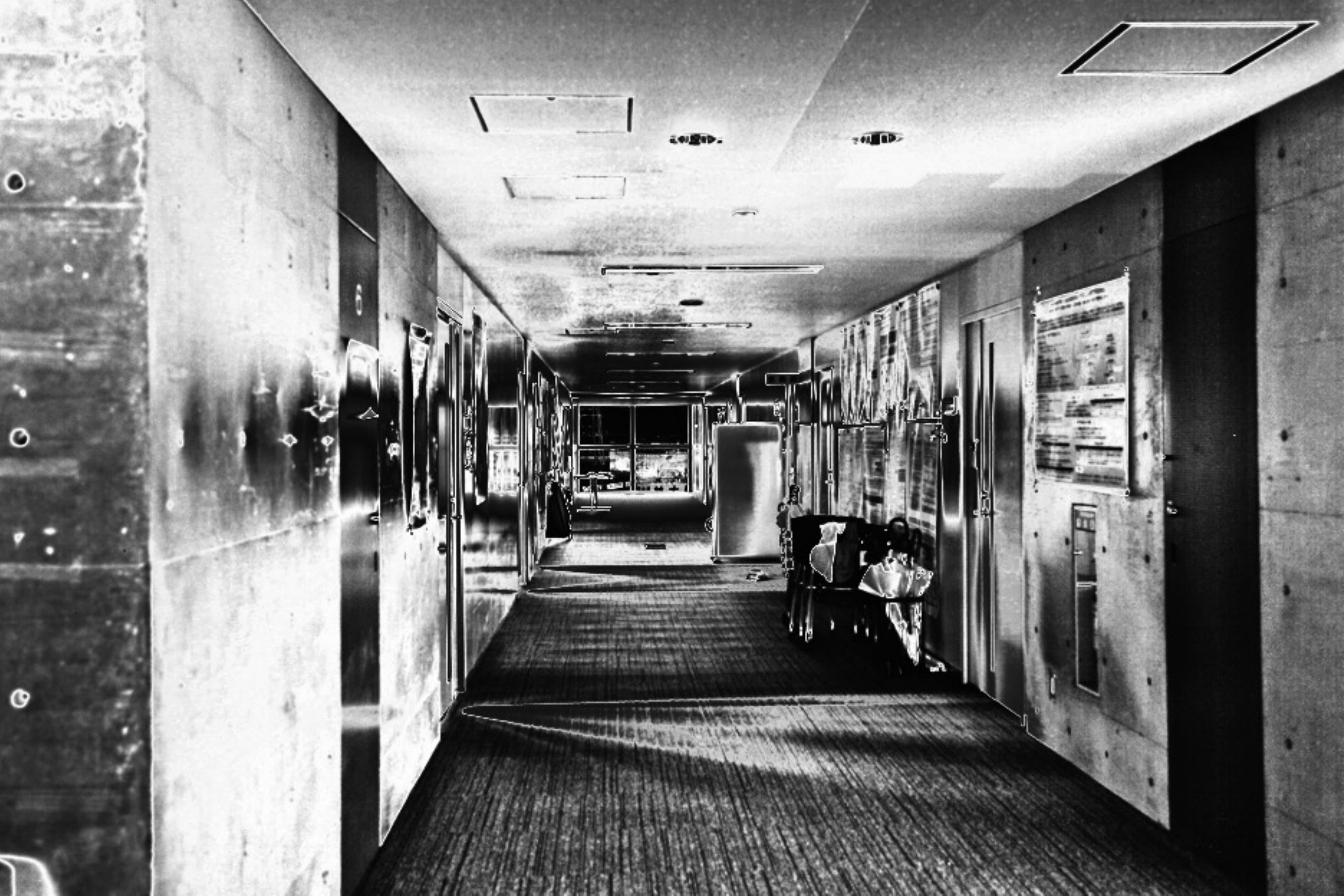}
  }\\
  \caption{Comparison among luminance adjustment methods
    under the use of Nejati's fusion method \cite{nejati2017fast}.
    Top to bottom: fused image $I_f$, zoom-in view of $I_f$, and well-exposedness map
    for adjusted multi-exposure images $I''_i$.
    For well-exposedness map, a brighter pixel indicates that the pixel is well exposed.}
  \label{fig:adjustment_difference}
\end{figure}

  Table \ref{tab:score_simulation2}
  summarizes average scores for 12 input images in terms of
  statistical naturalness and discrete entropy,
  and the second column, ``Input image'', shows average scores
  calculated by using input images having $0 \mathrm{[EV]}$.
  For each score (statistical naturalness $\in [0, 1]$
  and discrete entropy $\in [0, 8]$),
  a larger value means higher quality.
  The results indicate that the proposed method improves the quality of the fused images.
  It is also confirmed by comparing Fig. \ref{fig:adjustment_difference}(a)
  with Fig. \ref{fig:adjustment_difference}(b).
  Figure \ref{fig:adjustment_difference} also shows that the proposed method can
  keep the details in bright areas, and can enhance the details in dark areas.
  
\begin{figure}[!t]
  \centering
  \subfloat{
    \includegraphics[width=0.30\hsize]{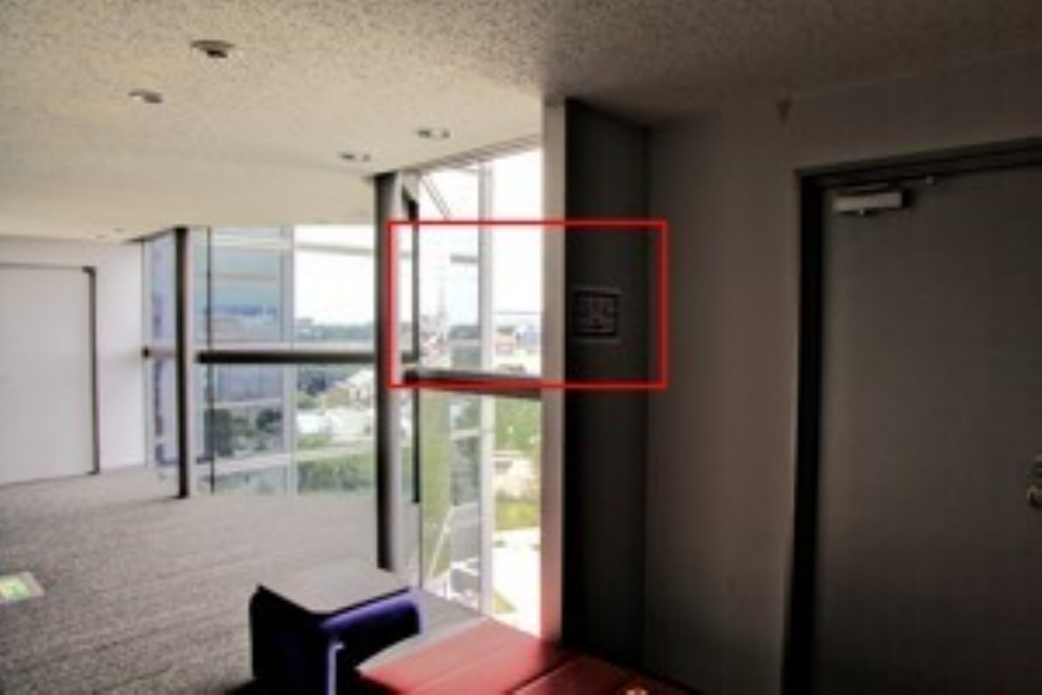}
  }
  \subfloat{
    \includegraphics[width=0.30\hsize]{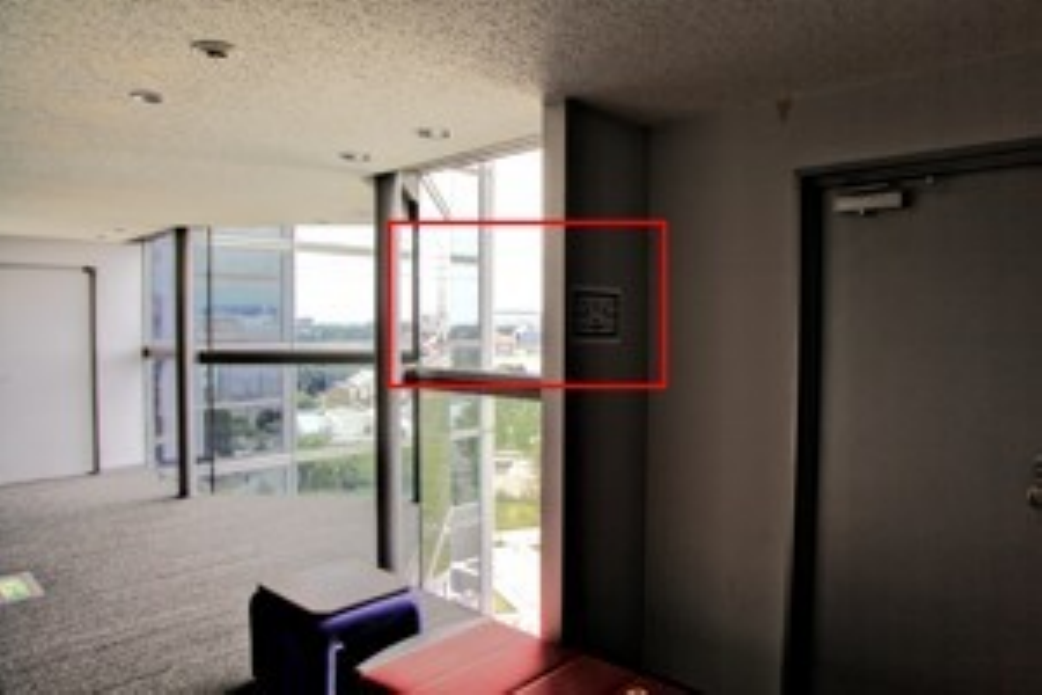}
  }
  \subfloat{
    \includegraphics[width=0.30\hsize]{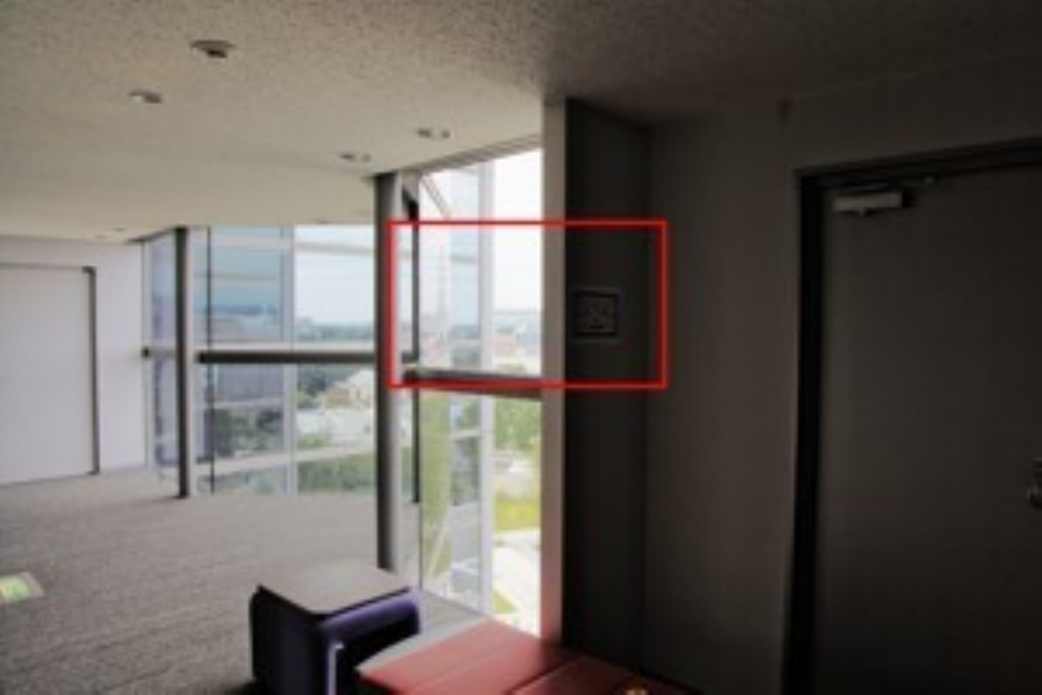}
  }\\
  \addtocounter{subfigure}{-3}
  \subfloat{
    \includegraphics[width=0.30\hsize]{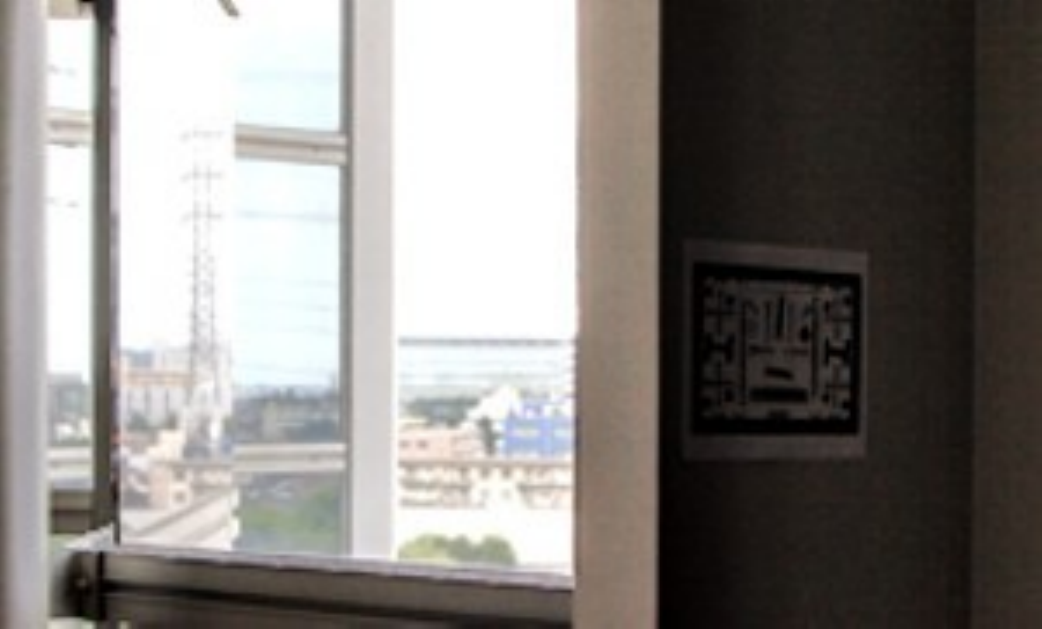}
  }
  \subfloat{
    \includegraphics[width=0.30\hsize]{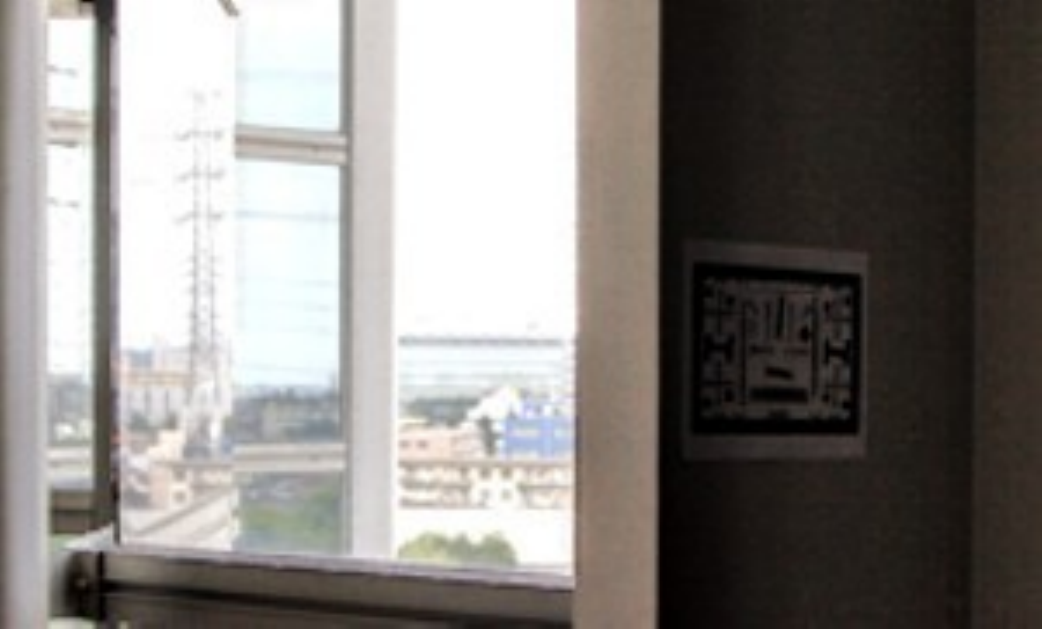}
  }
  \subfloat{
    \includegraphics[width=0.30\hsize]{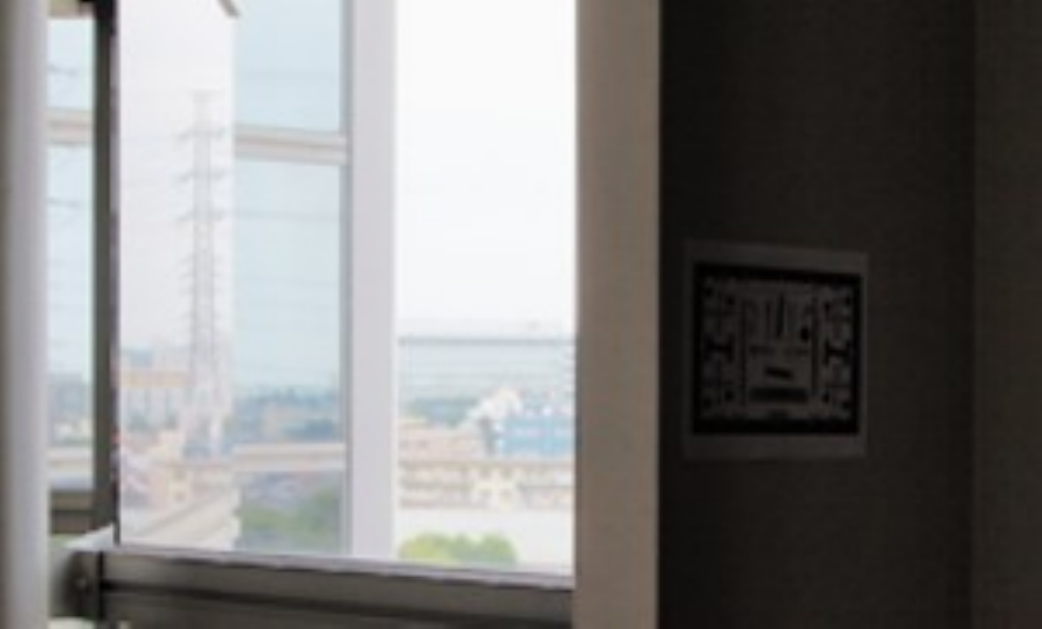}
  }\\
  \addtocounter{subfigure}{-3}
  \subfloat{
    \includegraphics[width=0.30\hsize]{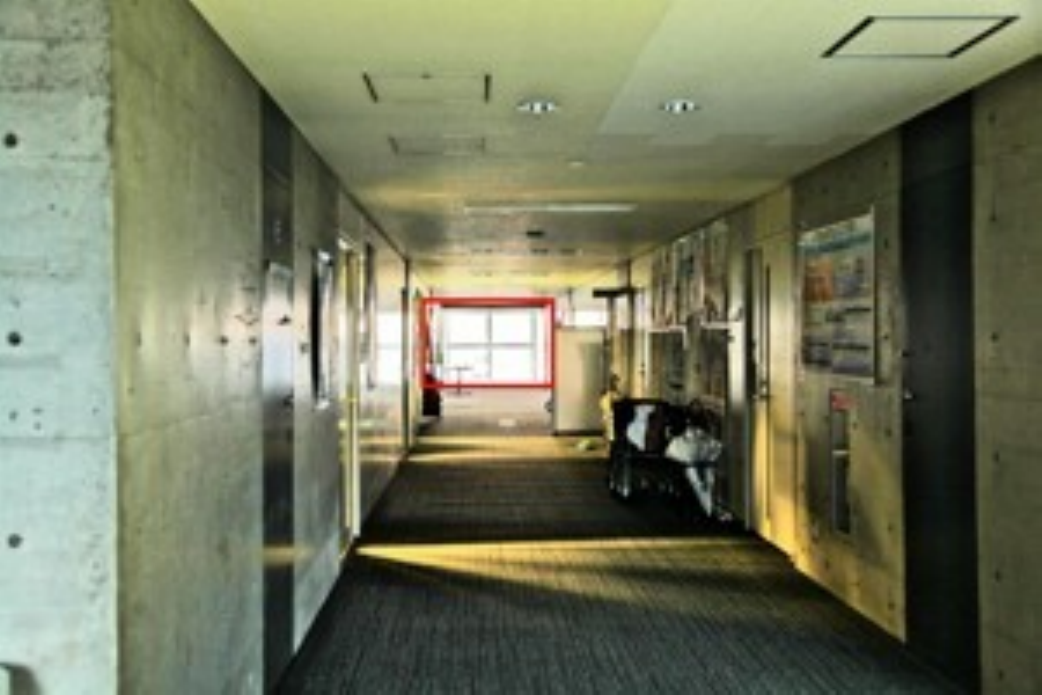}
  }
  \subfloat{
    \includegraphics[width=0.30\hsize]{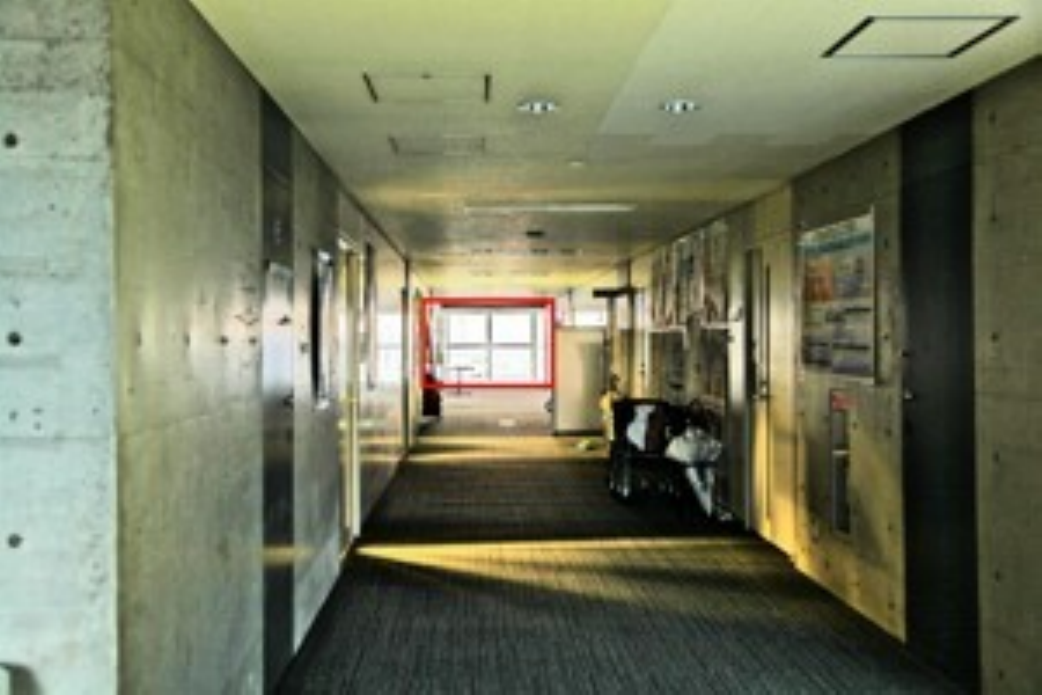}
  }
  \subfloat{
    \includegraphics[width=0.30\hsize]{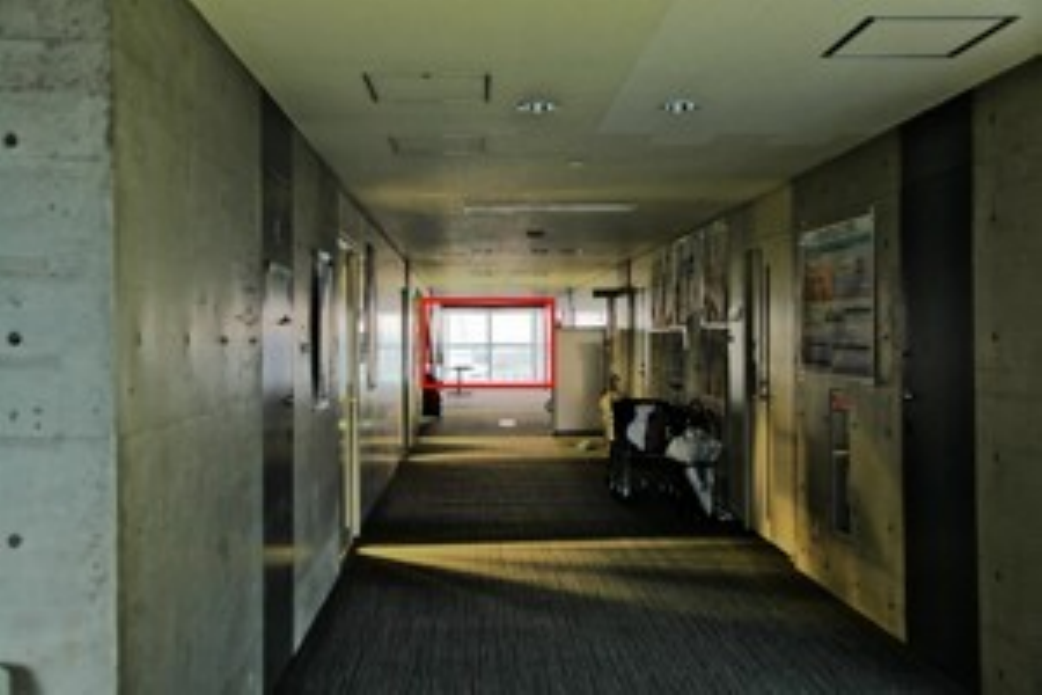}
  }\\
  \addtocounter{subfigure}{-3}
  \subfloat[Mertens \cite{mertens2009exposure}]{
    \includegraphics[width=0.30\hsize]{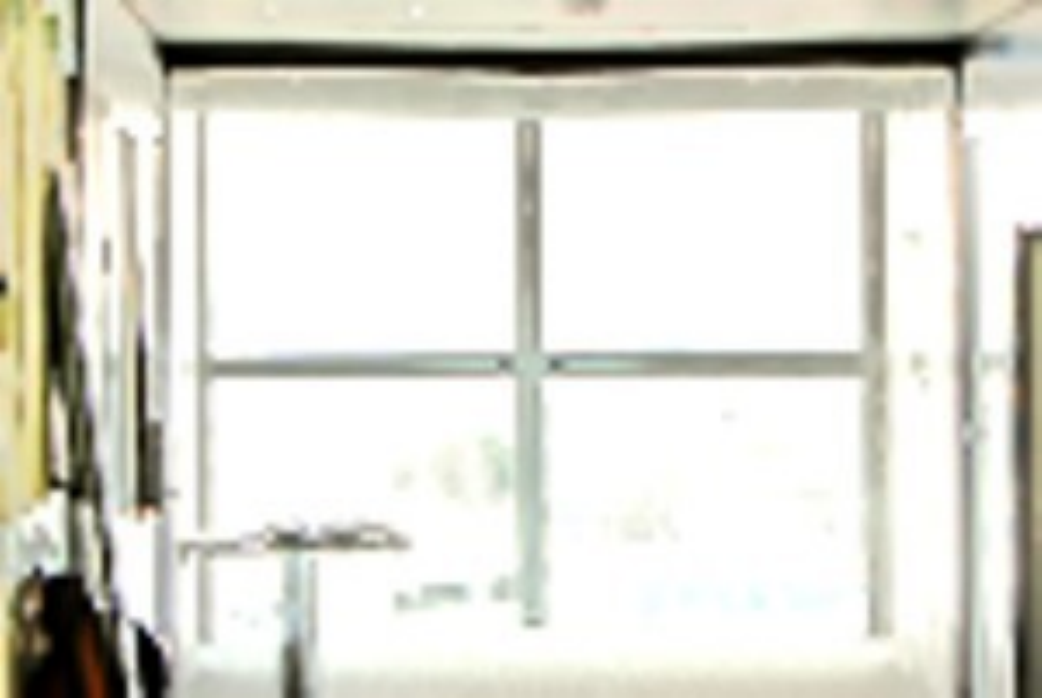}
  }
  \subfloat[Sakai \cite{sakai2015hybrid}]{
    \includegraphics[width=0.30\hsize]{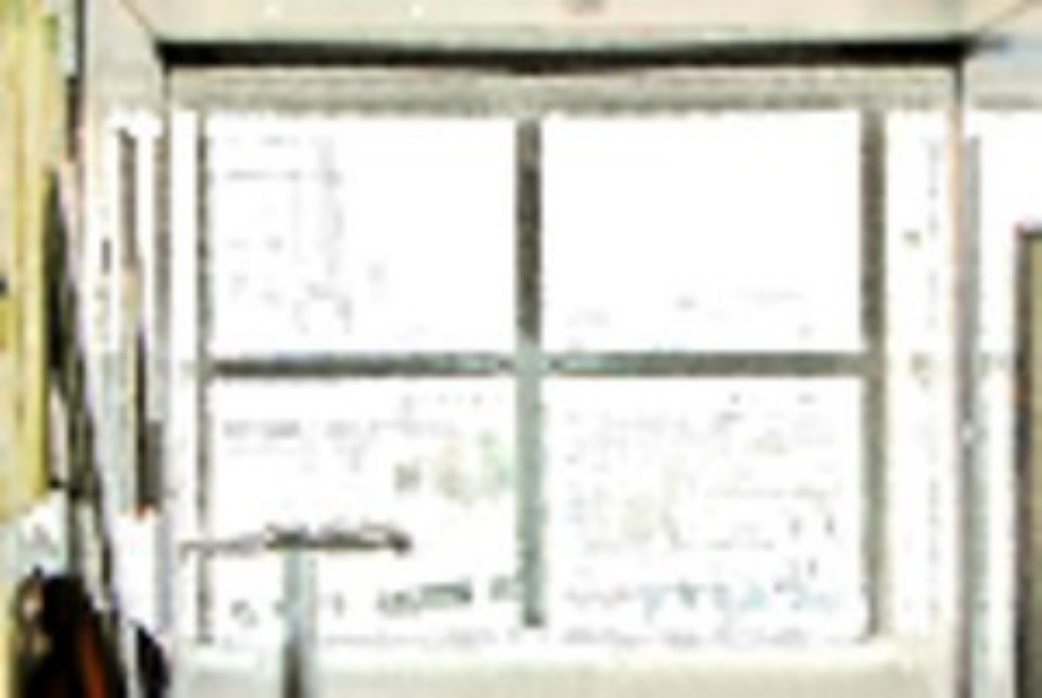}
  }
  \subfloat[Simple average]{
    \includegraphics[width=0.30\hsize]{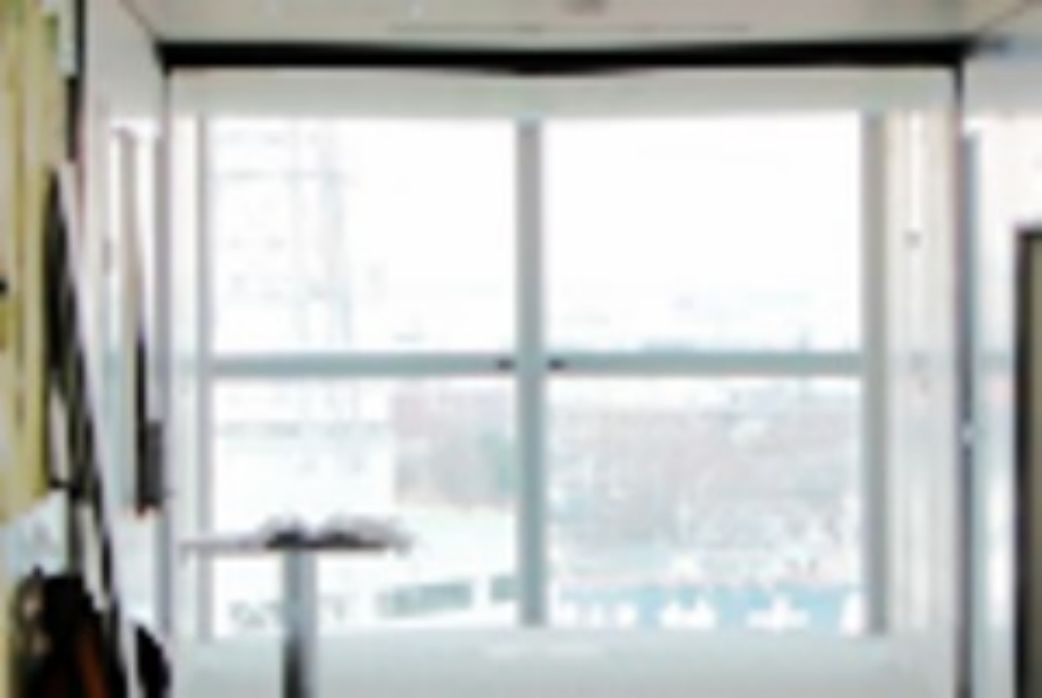}
  }\\
  \caption{Comparison among multi-exposure fusion methods under the use of the proposed method}
  \label{fig:fused_images}
\end{figure}

  From Table \ref{tab:score_simulation2}, it is confirmed that
  the simple average ($w_i(p) = 1$) under the use of the proposed adjustment method
  provided the highest score of each metric in all methods,
  but that without the adjustment brought the worst score.
  Figure \ref{fig:fused_images} denotes that images fused by the simple average with
  the proposed method
  represent bright areas with better quality than ones fused by the conventional methods.
  Hence, the proposed method can produce high quality images
  even when simple weights are used in eq. (\ref{eq:weighted_average}).
  
  For these reasons, 
  it is confirmed that the luminance adjustment is effective for multi-exposure image fusion.
  In addition, the use of the proposed luminance adjustment method
  is useful to produce high quality images which represent
  both bright and dark areas.
  Moreover, the proposed method enables us to utilize simple weights
  for multi-exposure image fusion, while keeping the quality of fused images.
%
%
\begin{table}[!t]
  \centering
  \caption{Experimental results.
  Boldface indicates the higher score.
  ``Without'' means that images $I_f$ are produced without luminance adjustment}
  \scalebox{0.90}{\footnotesize
    \begin{tabular}{l|c|cc|cc|cc|cc}\hline\hline
      \multirow{2}{*}{Method} & Input & \multicolumn{2}{|c|}{Mertens [5]}
        & \multicolumn{2}{|c|}{Sakai [9]}
        & \multicolumn{2}{|c|}{Nejati [10]}
        & \multicolumn{2}{|c}{Simple average} \\ \cdashline{3-10}
       & image & Without & Ours & Without & Ours
        & Without & Ours & Without & Ours \\ \hdashline
      Statistical & \multirow{2}{*}{0.092} & \multirow{2}{*}{0.083}
        & \multirow{2}{*}{\textbf{0.179}} & \multirow{2}{*}{0.079}
        & \multirow{2}{*}{\textbf{0.173}} & \multirow{2}{*}{0.089}
        & \multirow{2}{*}{\textbf{0.179}} & \multirow{2}{*}{0.045}
        & \multirow{2}{*}{\textbf{0.278}} \\
      Naturalness &  &  &  &  &  &  &  &  &  \\
      Discrete & \multirow{2}{*}{5.199} & \multirow{2}{*}{6.259}
        & \multirow{2}{*}{\textbf{6.952}} & \multirow{2}{*}{6.270}
        & \multirow{2}{*}{\textbf{6.968}} & \multirow{2}{*}{6.232}
        & \multirow{2}{*}{\textbf{6.920}} & \multirow{2}{*}{5.601}
        & \multirow{2}{*}{\textbf{7.071}}\\
      entropy &  &  &  &  &  &  &  &  &  \\
      \hline
    \end{tabular}
  }
  \label{tab:score_simulation2}
\end{table}
\section{Conclusion}
  This paper has proposed a novel luminance adjustment method
  based on automatic exposure compensation for multi-exposure fusion.
  The proposed method automatically adjusts the luminance of input multi-exposure images
  to suitable ones for multi-exposure fusion.
  The proposed method also enables us to utilize simple weights
  for multi-exposure image fusion, while keeping the quality of fused images.
  Experimental results have showed
  the effectiveness of the luminance adjustment for multi-exposure image fusion
  in terms of the well-exposedness.
  Moreover, it has been confirmed that fusion methods can produce high quality images
  under the use of the proposed luminance adjustment method,
  in terms of statistical naturalness and discrete entropy.
%


\end{document}